\documentclass[letterpaper, 10 pt, confkrence]{ieeeconf}
\usepackage{amsmath}
\usepackage{mathtools}
\usepackage{algorithm}
\usepackage[noend]{algpseudocode}
\usepackage{hyperref}
\usepackage{multirow}

\usepackage{color,soul}

\usepackage{subcaption}
\usepackage{kantlipsum} %<- For dummy text
\usepackage{mwe} %<- For dummy images
\usepackage[squaren, Gray, cdot]{SIunits}
\usepackage{hhline}

\DeclareCaptionLabelSeparator{periodspace}{.\quad}
\IEEEoverridecommandlockouts                              
\title{\LARGE \bf Shape-aware Safe Corridors Generation using Voxel Grids}

\author{Charbel Toumieh and Alain Lambert% <-this % stops a space
\thanks{The two authors are with the Universit\'e Paris-Saclay, CNRS, Laboratoire Interdisciplinaire des Sciences du Numérique, 91405, Orsay, France \url{https://www.lisn.upsaclay.fr}}%
}
\begin{document}

\maketitle
\thispagestyle{empty}
\pagestyle{empty}

%%%%%%%%%%%%%%%%%%%%%%%%%%%%%%%%%%%%%%%%%%%%%%%%%%%%%%%%%%%%%%%%%%%%%%%%%%%%%%%%
\begin{abstract}
Safe Corridors (a series of overlapping convex shapes) have been used recently in multiple state-of-the-art motion planning methods. They allow to represent the free space in the environment in an efficient way for collision avoidance. In this paper, we propose a new framework for generating Safe Corridors. We assume that we have a voxel grid representation of the environment. The proposed framework improves on a previous state-of-the-art voxel grid based Safe Corridor generation method. It also creates a connectivity graph between polyhedra of a given Safe Corridor that allows to know which polyhedra intersect with each other. The connectivity graph can be used in planning methods to reduce computation time. The method is compared to other state-of-the-art methods in simulations in terms of computation time, volume covered, safety, number of polyhedron per Safe Corridor and number of constraints per polyhedron.  
\end{abstract}

%%%%%%%%%%%%%%%%%%%%%%%%%%%%%%%%%%%%%%%%%%%%%%%%%%%%%%%%%%%%%%%%%%%%%%%%%%%%%%%%
\section{INTRODUCTION}
\subsection{Problem statement}
Safe Corridors (SC) have been used extensively in robot planning frameworks \cite{toumieh2020planning}, \cite{tordesillas2020faster}, \cite{toumieh2022time}, \cite{toumieh2022multi}, \cite{park2020efficient} \cite{toumieh2022mace} \cite{toumieh2022dyn}. They offer a representation of the free space in the environment that a robot can use to plan in. A method that is low compute and that generates high quality Safe Corridors is thus of great interest.

\subsection{Related work} \label{sect:related_work}
Many works in the literature addressed the problem of creating Safe Corridors in free space for robot planning.

In \cite{7487283}, the authors use an OctoMap \cite{hornung2013octomap} and create a sequence
of overlapping axes-aligned cubes. This method is computationally efficient and well-suited for embedded systems. However, it is  non-generic and only works well when the obstacles are rectangular parallelopipeds.

In \cite{gao2019flying}, a pointcloud is used by the authors as a representation of the environment. The pointcloud is used to partition free space into overlapping spheres. This approach can result in a large number of spheres in a low complexity environment (a narrow straight corridor). 

In \cite{deits2015computing}, the authors search for an ellipsoid and a group of hyperplanes (convex polyhedron) that separate it from the obstacles (which are represented as convex polyhedra). The ellipsoid and polyhedron are found by solving an optimization problem (nonconvex) that maximizes the volume of the ellipsoid.  
This method is considered generic: there is not any limitations to the shape of the generated convex polyhedron in the sense that it is a sphere or a cube. However it has a high computation time that makes it unsuitable for real-time applications.

In \cite{liu2017planning}, a pointcloud representation of the environment is used. The point cloud is downsampled before it is used by the algorithm \cite{tordesillas2020faster}. Otherwise, the computation time becomes too high for real-time applications. The method inflates an ellipsoid around a line/seed point until the ellipsoid hits an obstacle point. Then, the authors generate a tangent plane to the ellipsoid at the obstacle point (hyperplane), and remove obstacle points that are on the side of the hyperplane that doesn't contain the seed. The ellipsoid is then further inflated until one of the remaining obstacle points is hit. The authors then generate a new hyperplane and remove a portion of the obstacle points in the same manner. The authors repeat this process until no obstacle points remain in the pointcloud. However, this approach can result in unsafe \textit{Safe} Corridors that intersect with the obstacles by penetrating between the downsampled points.

Finally, in \cite{toumieh2020convex}, the authors use an occupancy grid representation of the environment. The authors start with a seed voxel (around which they want to find a convex polyhedron). This seed voxel is then expanded in all directions until it no longer can be expanded according to some rules that are defined in \cite{toumieh2020convex}. The entity that is expanded is what the authors call a convex grid (a group of voxel cells), in which a convex polyhedron is inscribed. As the convex grid is expanded, the polyhedron (which is inscribed) is modified and limits how the convex grid is modified/expanded at the next iteration. The authors choose multiple seeds along a path and expands them into convex grids/polyhedra to create a Safe Corridor. The advantage of this method over \cite{liu2017planning} is the fact that it guarantees safety of the Safe Corridor even when dealing with downsampled point clouds since by construction, the generated polyhedra have no intersection with the occupied voxels of the occupancy grid.

\subsection{Contribution}
Our method is an extension of the method presented in \cite{toumieh2020convex}. The main additions are the following:

\begin{itemize}
    \item The addition of new conditions for expanding a border which makes our method have a better decomposition of the free space.
    \item The creation of a connectivity graph using the convex grid/polyhedron duality that allows to know which polyhedra intersect with each others.
\end{itemize}

We compare our method to \cite{liu2017planning} and \cite{toumieh2020convex} in terms of \textbf{genericness} (number of polyhedra in the SC), \textbf{volume} covered by the SC, \textbf{computation time} and \textbf{number of constraints} that describe each polyhedron of the SC. A more detailed explanation of these performance metrics can be found in \cite{toumieh2020convex}.

\section{The method}
The method is an extension of the work done in \cite{toumieh2020convex}. We will briefly describe the method used in \cite{toumieh2020convex} then mention the modifications we made to make it better.

\begin{figure*}
\begin{subfigure}{0.33\textwidth}
\centering
\includegraphics[trim={0cm 0cm 0cm 0cm},clip,width=1\linewidth]{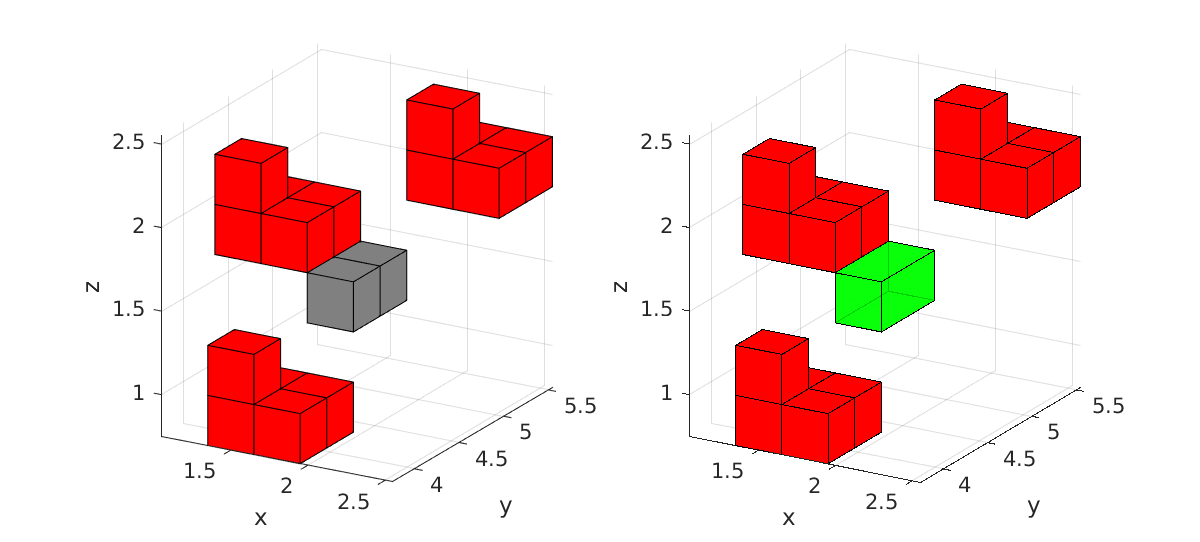}
\caption{k = 1}
\end{subfigure}
\begin{subfigure}{0.33\textwidth}
\centering
\includegraphics[trim={0cm 0cm 0cm 0cm},clip,width=1\linewidth]{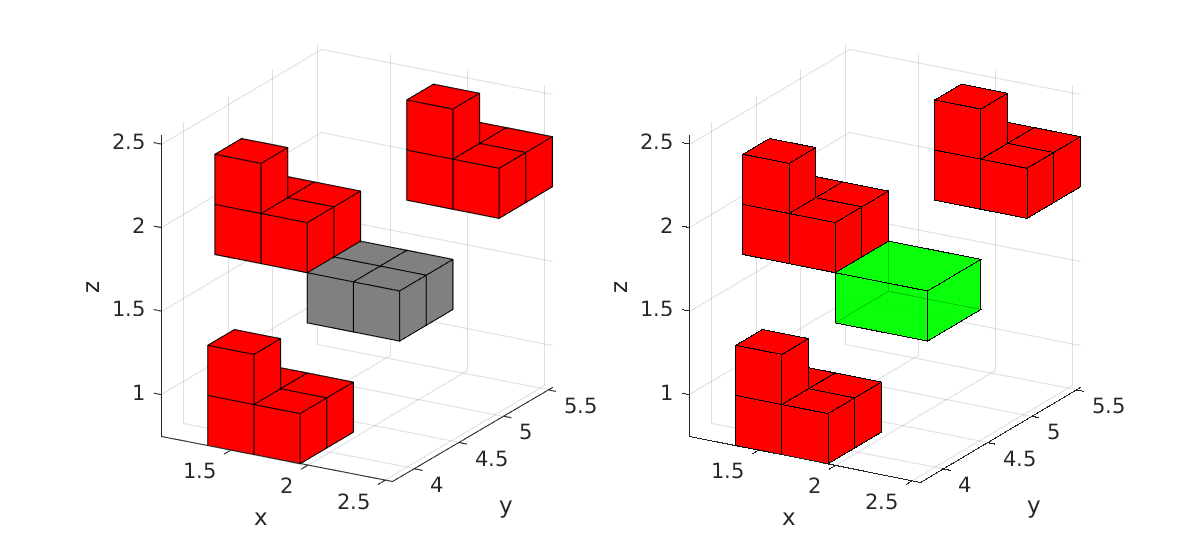}
\caption{k = 2}
\end{subfigure}
\begin{subfigure}{0.33\textwidth}
\centering
\includegraphics[trim={0cm 0cm 0cm 0cm},clip,width=1\linewidth]{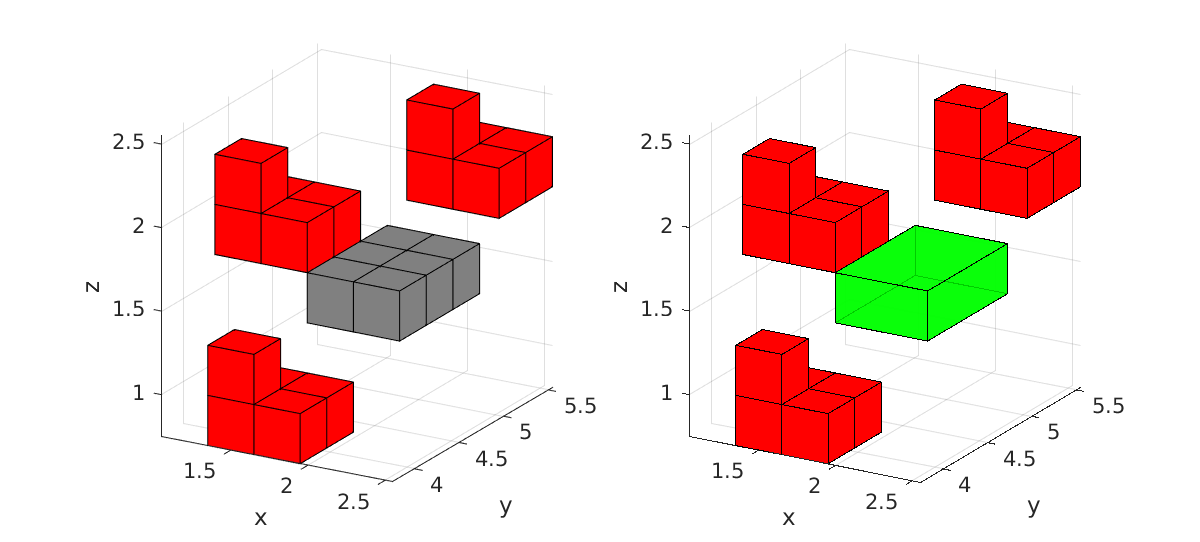}
\caption{k = 3}
\end{subfigure}

\begin{subfigure}{0.33\textwidth}
\centering
\includegraphics[trim={0cm 0cm 0cm 0cm},clip,width=1\linewidth]{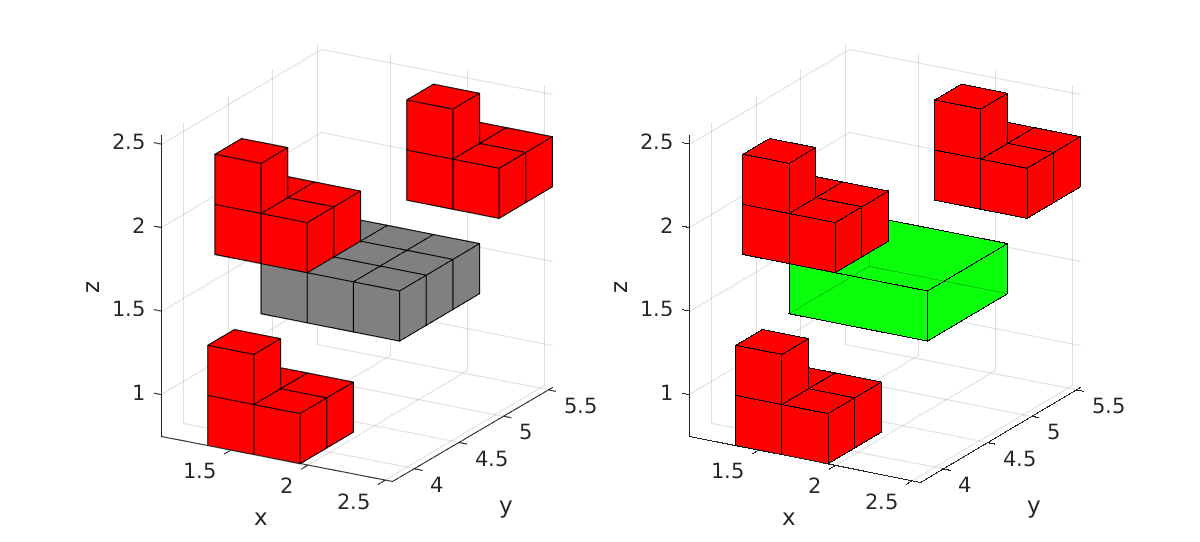}
\caption{k = 4}
\end{subfigure}
\begin{subfigure}{0.33\textwidth}
\centering
\includegraphics[trim={0cm 0cm 0cm 0cm},clip,width=1\linewidth]{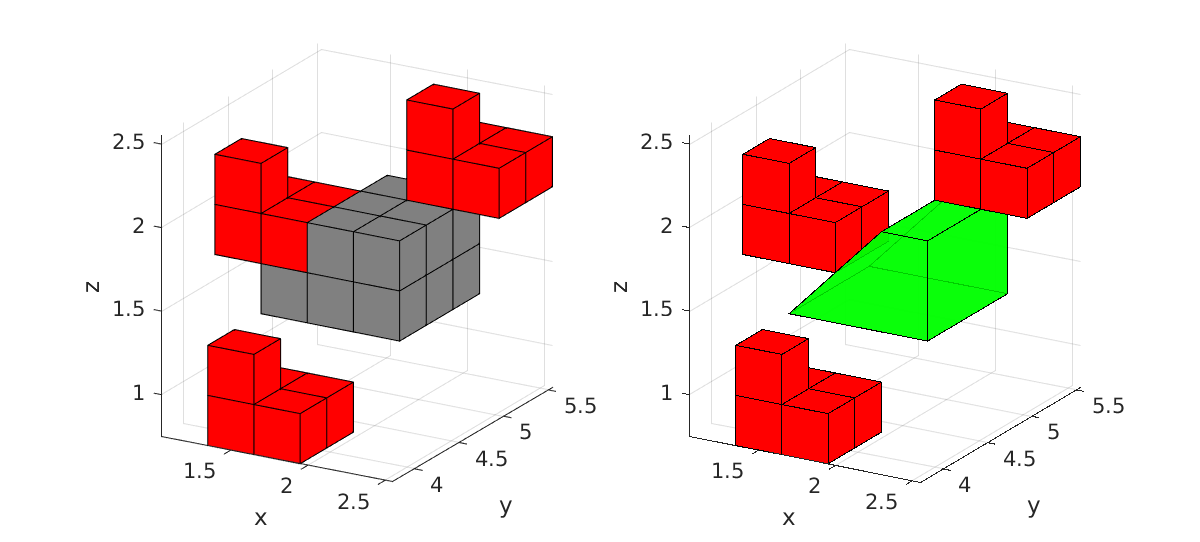}
\caption{k = 5}
\end{subfigure}
\begin{subfigure}{0.33\textwidth}
\centering
\includegraphics[trim={0cm 0cm 0cm 0cm},clip,width=1\linewidth]{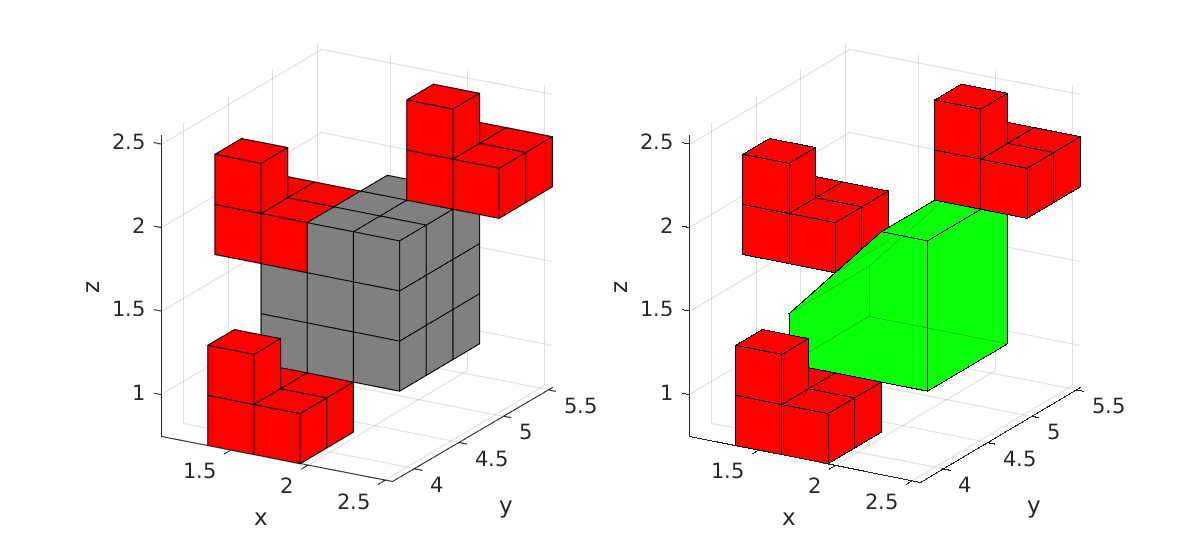}
\caption{k = 6}
\end{subfigure}

\begin{subfigure}{0.33\textwidth}
\centering
\includegraphics[trim={0cm 0cm 0cm 0cm},clip,width=1\linewidth]{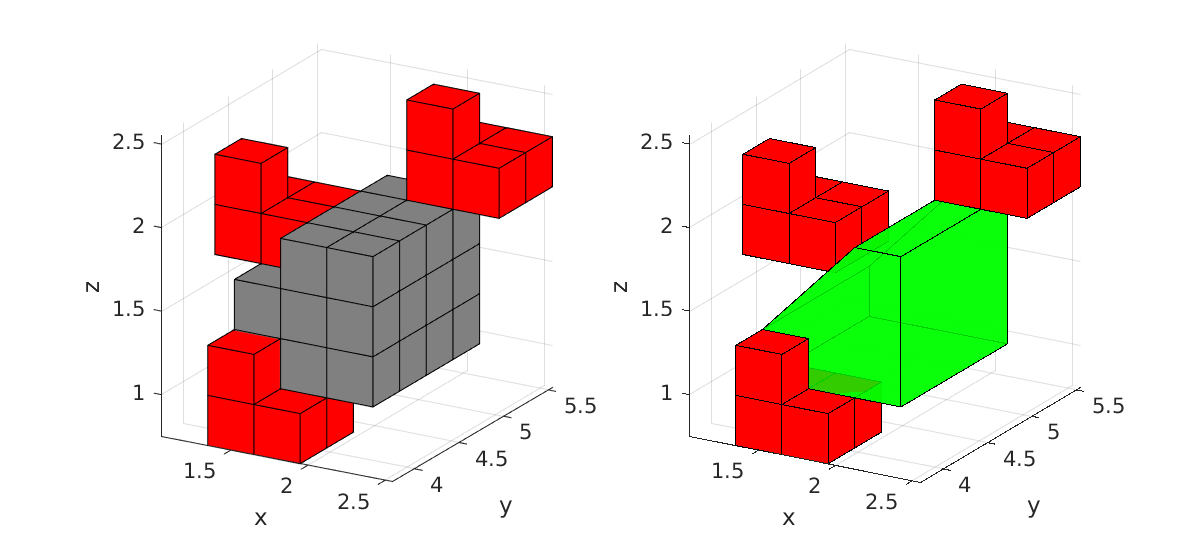}
\caption{k = 7}
\end{subfigure}
\begin{subfigure}{0.33\textwidth}
\centering
\includegraphics[trim={0cm 0cm 0cm 0cm},clip,width=1\linewidth]{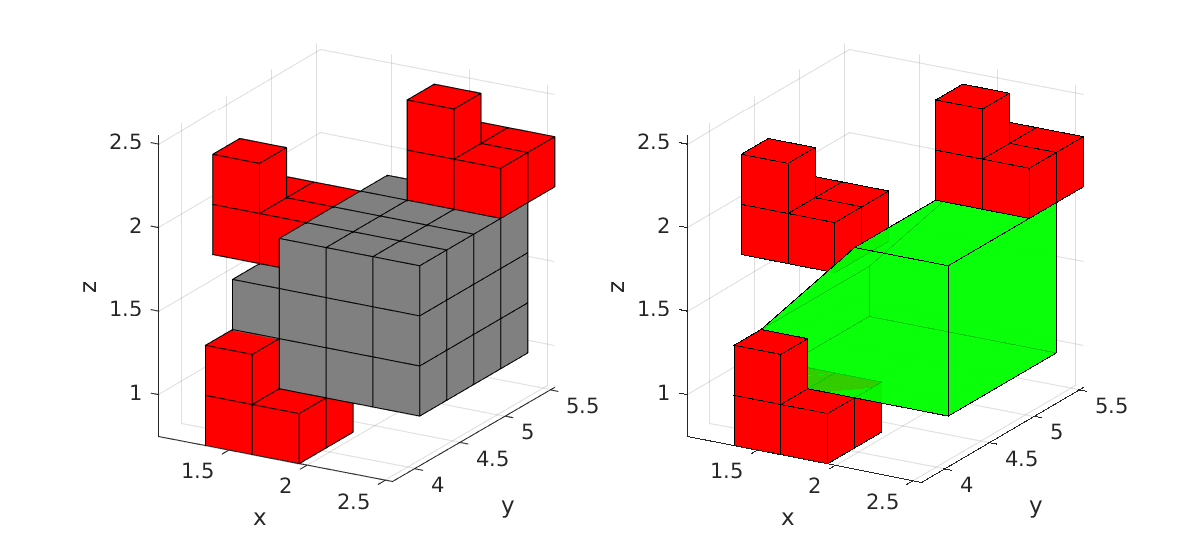}
\caption{k = 8}
\end{subfigure}
\begin{subfigure}{0.33\textwidth}
\centering
\includegraphics[trim={0cm 0cm 0cm 0cm},clip,width=1\linewidth]{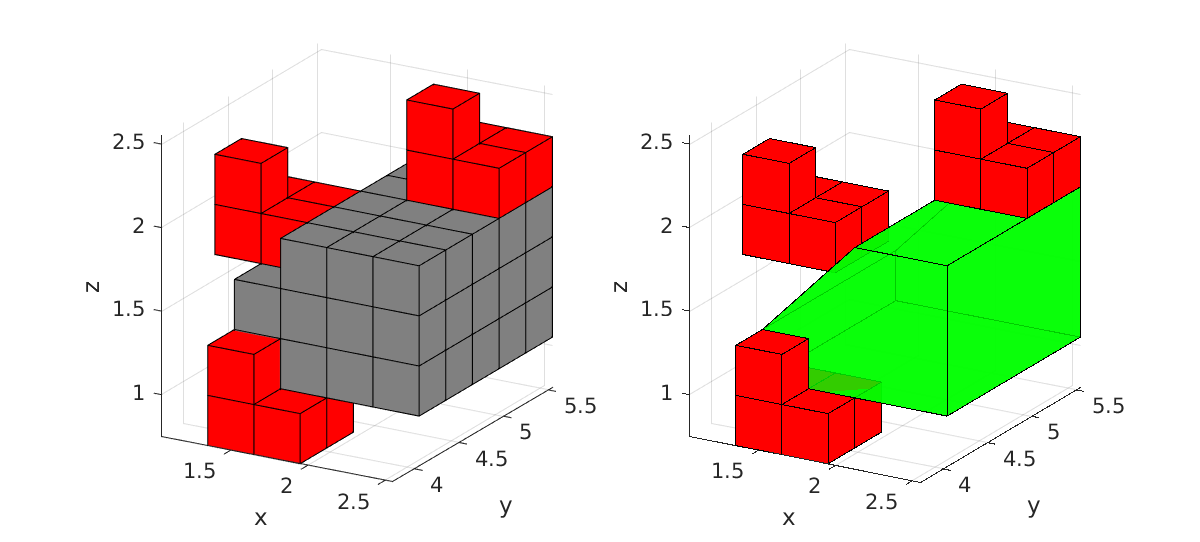}
\caption{k = 9}
\end{subfigure}

\begin{subfigure}{0.33\textwidth}
\centering
\includegraphics[trim={0cm 0cm 0cm 0cm},clip,width=1\linewidth]{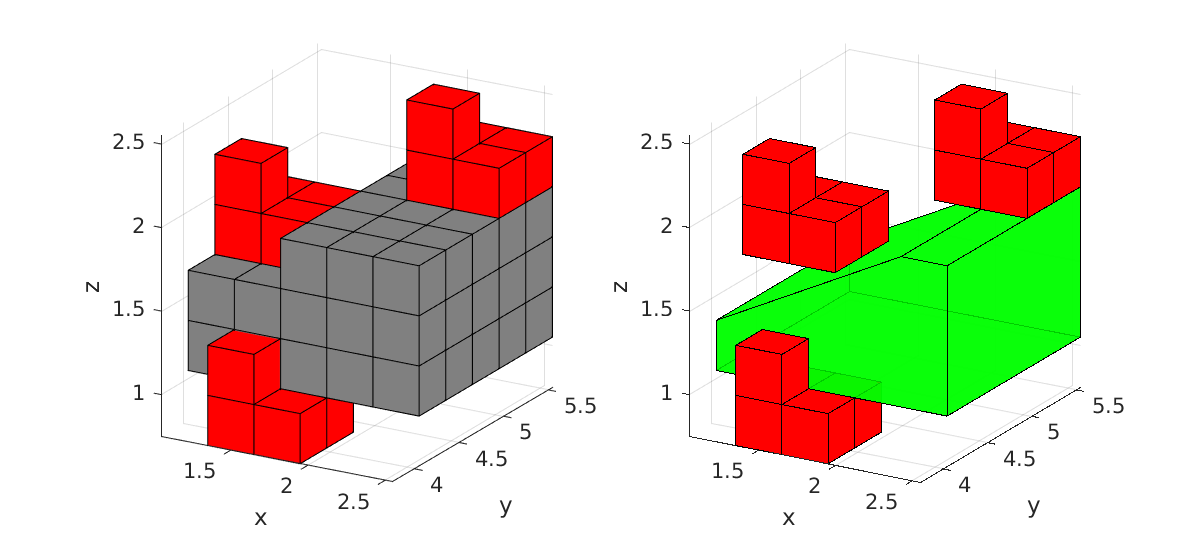}
\caption{k = 10}
\end{subfigure}
\begin{subfigure}{0.33\textwidth}
\centering
\includegraphics[trim={0cm 0cm 0cm 0cm},clip,width=1\linewidth]{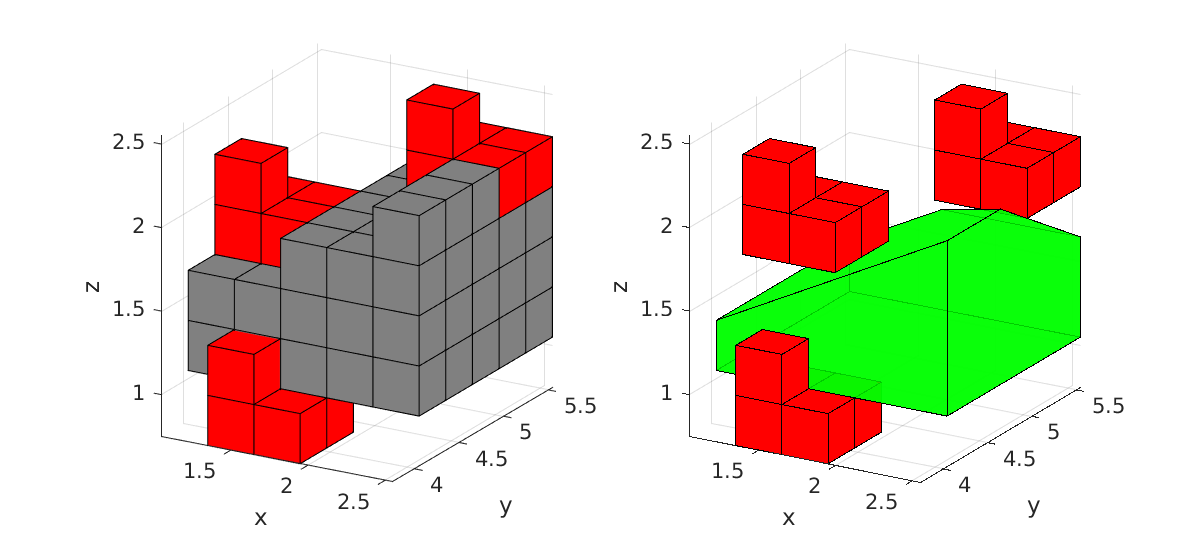}
\caption{k = 11}
\end{subfigure}
\begin{subfigure}{0.33\textwidth}
\centering
\includegraphics[trim={0cm 0cm 0cm 0cm},clip,width=1\linewidth]{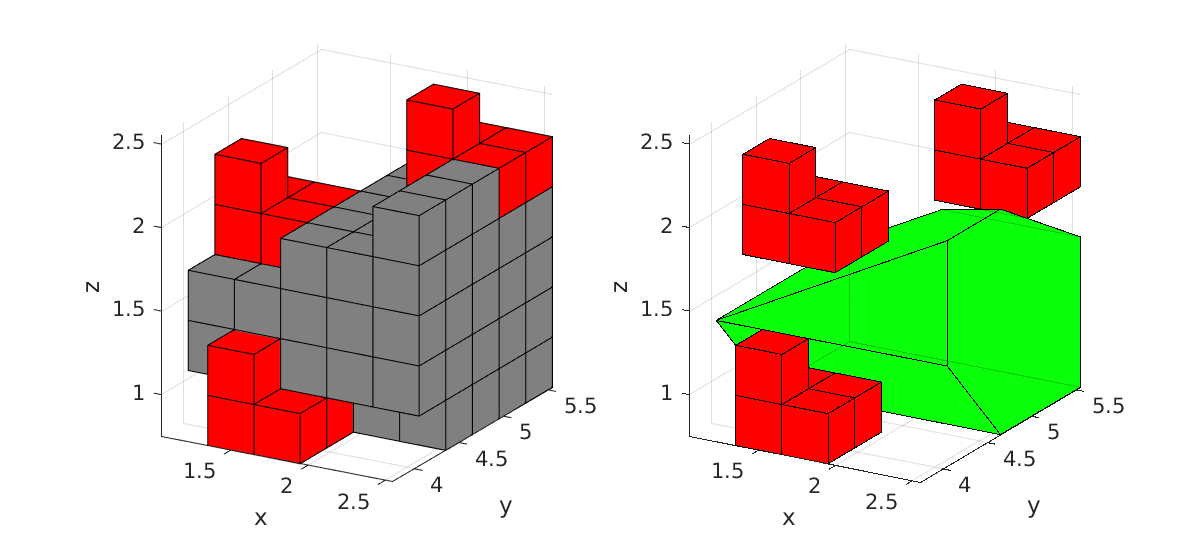}
\caption{k = 12}
\end{subfigure}
\caption{We show the evolution for 12 iterations of the \textbf{convex grid} (in \textbf{grey}) and its corresponding inscribed polyhedron (in \textbf{green}). The \textbf{obstacles} are represented as \textbf{red} voxels.}
\label{fig:seed_decomp}
\end{figure*}

We start with a seed voxel around which we want to find a convex polyhedron. The seed is then expanded in all directions in a cyclic manner until we no longer can according to the rules defined in \cite{toumieh2020convex} or the maximum number of expansions is reached. The entity that we expand is what the authors of \cite{toumieh2020convex} call a convex grid, in which a convex polyhedron is inscribed. Every time we expand in a direction, the inscribed polyhedron is changed and affects how the convex grid is expanded at the next iterations. A visualisation of the expansion process for 12 iterations is shown in Fig. \ref{fig:seed_decomp}. For more details into how the expansion algorithm works as well as the rules that govern it, we refer the reader to \cite{toumieh2020convex}.

\subsection{Polyhedra volume heuristic}
\begin{figure*}
\begin{subfigure}{1\textwidth}
\centering
\includegraphics[trim={0cm 0cm 0cm 0cm},clip,width=0.95\linewidth]{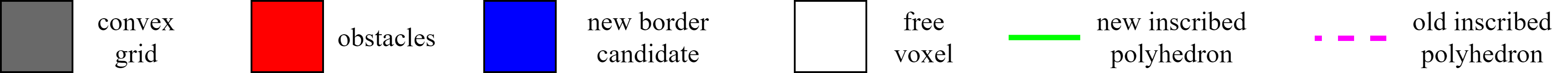}
\caption{Legend}
\label{fig:legend}
\end{subfigure}

\begin{subfigure}{0.5\textwidth}
\centering
\includegraphics[trim={0cm 0cm 0cm -0.5cm},clip,width=0.8\linewidth]{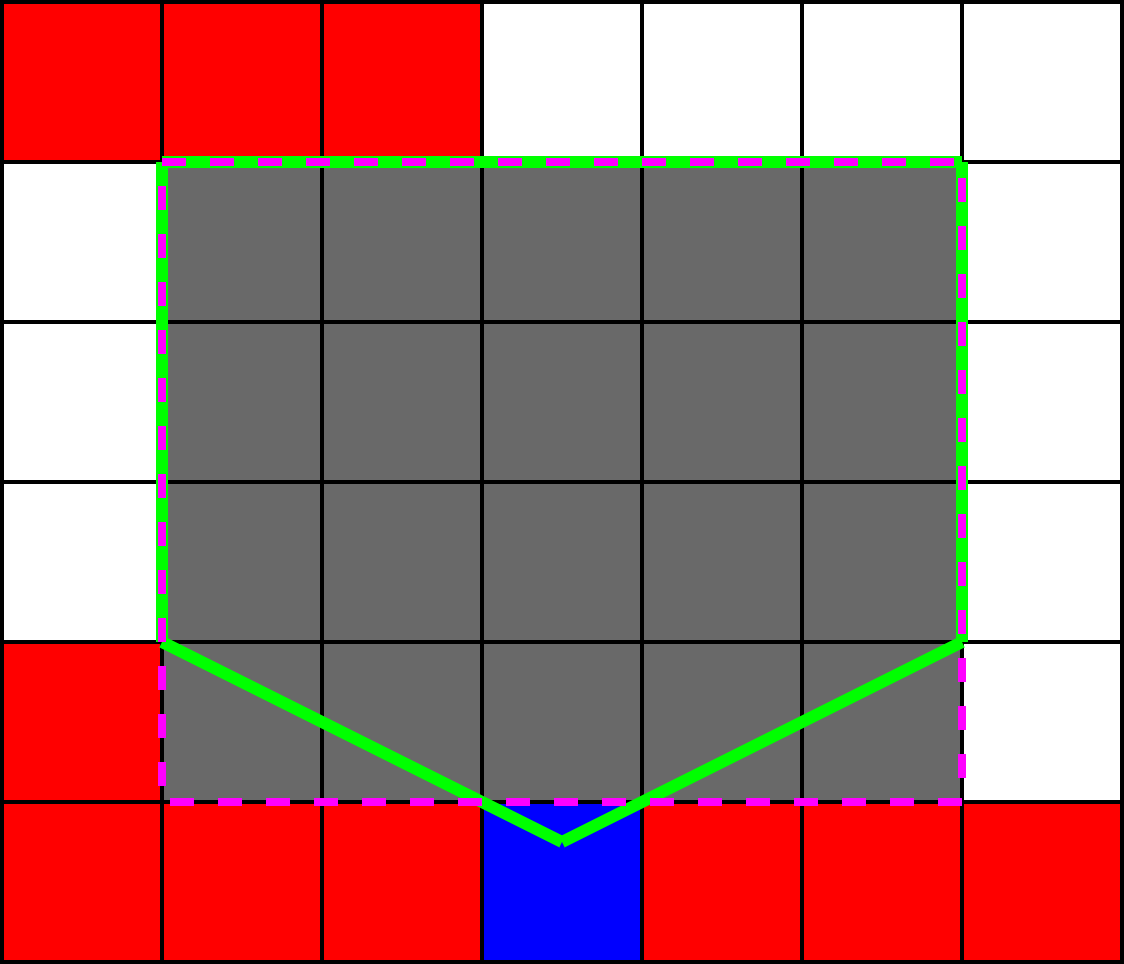}
\caption{Invalid expansion}
\label{fig:invalid_exp}
\end{subfigure}
\begin{subfigure}{0.5\textwidth}
\centering
\includegraphics[trim={0cm 0cm 0cm -0.5cm},clip,width=0.8\linewidth]{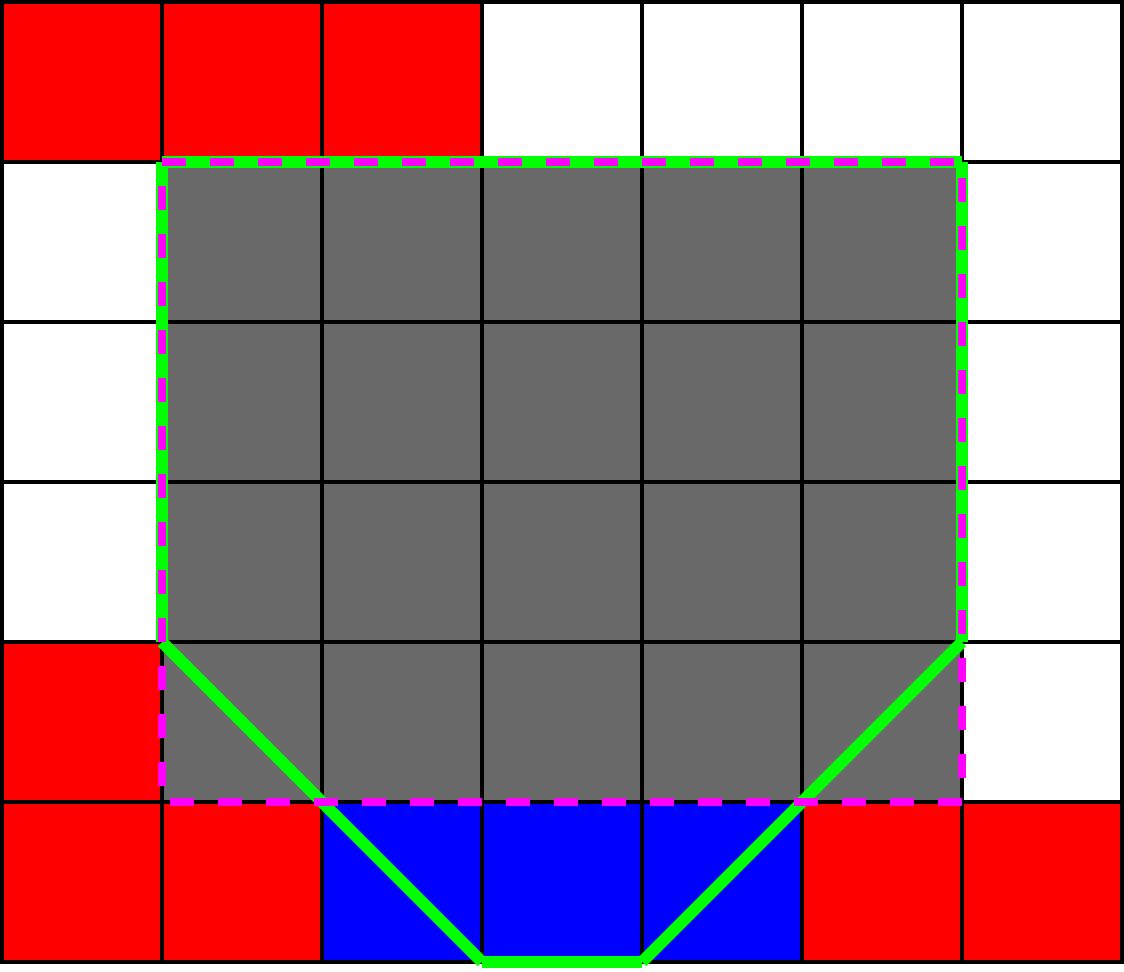}
\caption{Valid expansion}
\label{fig:valid_exp}
\end{subfigure}

\caption{We show a 2D example of the heuristic that invalidates a border expansion. The heuristic that we use is that the number of the new border candidate should be bigger then the border that we expand in terms of number of voxels. In the first case \ref{fig:invalid_exp}, the new border candidate that results from the expansion has 1 voxel  whereas the number of voxels that we expanded was 5. In this case the expansion is invalid according to the heuristic. One can see that if we validate it, the new inscribed polyhedron (in green) in the convex grid covers less space then the old inscribed polyhedron (in dashed purple). A valid expansion according to the heuristic is shown in \ref{fig:valid_exp}, where the new border candidate has 3 voxels (more than half the voxels expanded - in this case 2.5). One can see that the new inscribed polyhedron covers more space than the old one.}
\label{fig:first_condition}
\end{figure*}

We take the rules of expansion defined in \cite{toumieh2020convex} and add the following condition: when we find a new border candidate, we only validate it if the number of voxels in it is bigger then half the number of voxels in the border we are expanding. This heuristic generally avoids making the volume covered by the inscribed polyhedron smaller after the expansion. An expansion example can be seen in Fig. \ref{fig:first_condition}.

The rules defined in \cite{toumieh2020convex} in addition to this condition is what we will refer to as legacy rules in the remainder of this paper to validate/invalidate an expansion.

\subsection{Newly initialized corners}
When a new border candidate requires the initialization of a new corner (see \cite{toumieh2020convex} for more details on when corners are initialized), we have 2 cases. The first case is when the direction of the corner is fixed (which we will denote \textbf{C1}), and the other case is when it is not fixed (which will we denote \textbf{C2}) - Fig. \ref{fig:corner_check}. Each case requires additional checks and rules to the legacy rules (applied to each newly initialized corner) before we validate the new border candidate.

\subsubsection{First check:} 
\begin{figure*}
\begin{subfigure}{0.5\textwidth}
\centering
\includegraphics[trim={0cm 0cm 0cm 0cm},clip,width=0.8\linewidth]{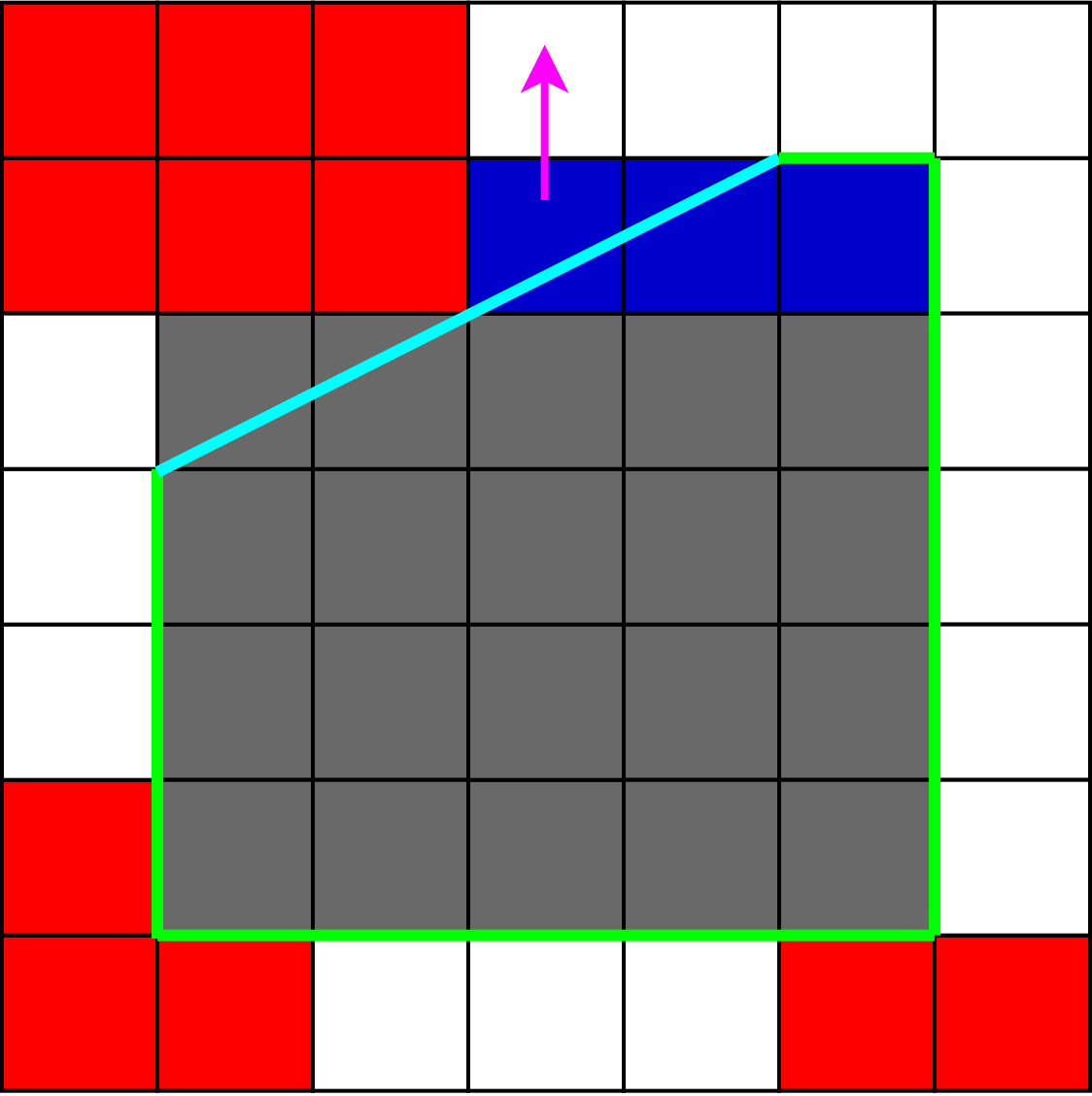}
\caption{Case \textbf{C1}: direction fixed}
\label{fig:corner_check_1}
\end{subfigure}
\begin{subfigure}{0.5\textwidth}
\centering
\includegraphics[trim={0cm 0cm 0cm 0cm},clip,width=0.8\linewidth]{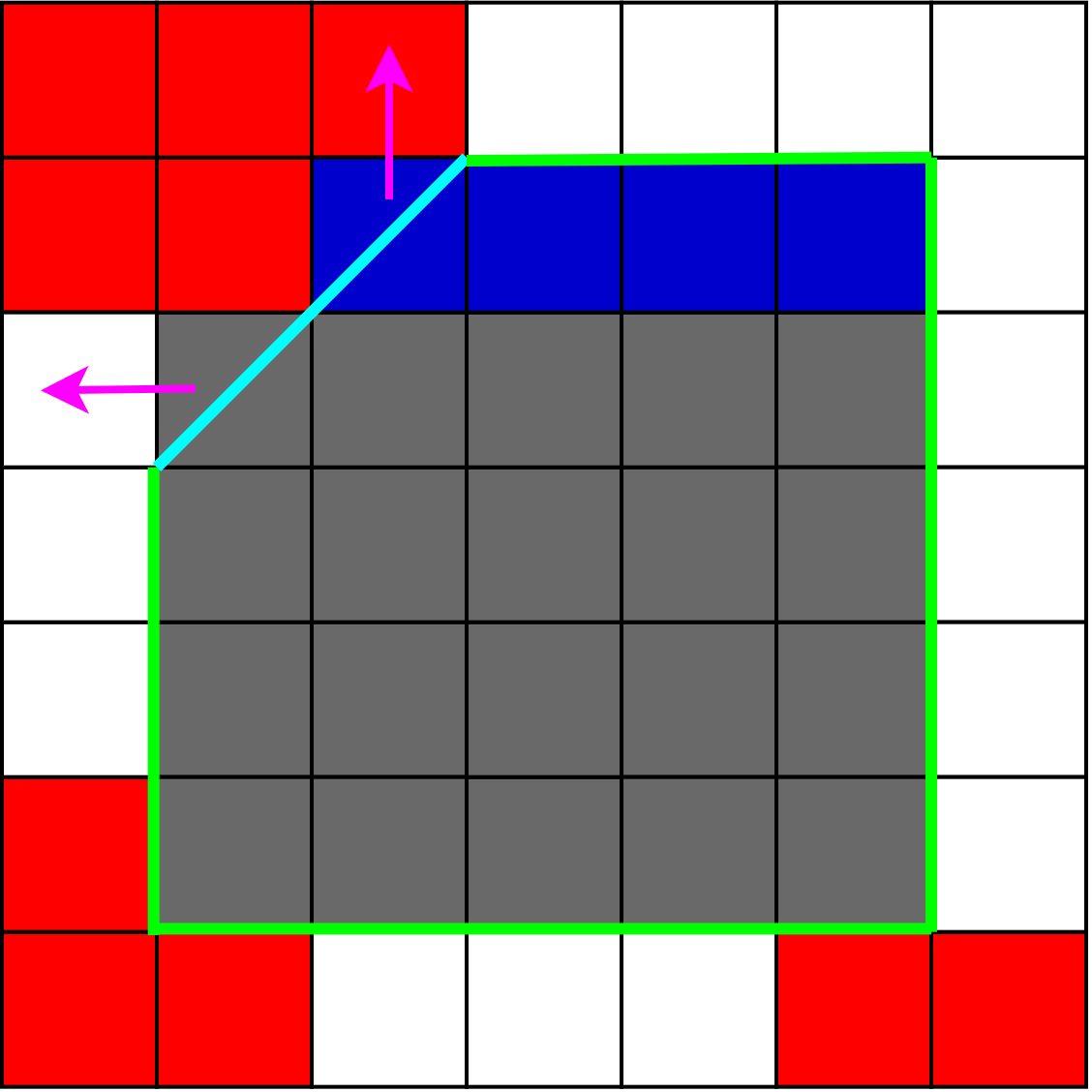}
\caption{Case \textbf{C2}: direction unknown}
\label{fig:corner_check_2}
\end{subfigure}
\caption{We show a 2D example of the 2 cases that are treated by our method. In the first cast \ref{fig:corner_check_1} the corner of the new inscribed polyhedron (shown in cyan) has a fixed direction according to the legacy rules, whereas in case 2 \ref{fig:corner_check_2} the corner direction is not determined yet. We also show with the magenta arrows the expansion of the voxels adjacent to corner in each case. These voxels are used to determine whether the expansion is valid in each case. In the case shown if \ref{fig:corner_check_1}, the expansion is not valid, whereas in \ref{fig:corner_check_2} the expansion is valid because expanding one of the voxels results in an occupied voxel.}
\label{fig:corner_check}
\end{figure*}

First we check the following condition for each of the newly initialized corners: we expand all the voxels of the new border candidate that are adjacent to the newly initialized corner in the same direction as the new border candidate. If all of the newly expanded voxels are empty and we are in the case \textbf{C1} then the expansion is invalid (Fig. \ref{fig:corner_check_1}). If they are empty and we are in \textbf{C2} then we expand the voxels in the initial expansion border that are adjacent to the newly initialized corner. If the newly expanded voxels are also empty then the expansion of the new border candidate is not valid (Fig. \ref{fig:corner_check_2}).

\subsubsection{Second check:}
\begin{figure*}
\begin{subfigure}{0.33\textwidth}
\centering
\includegraphics[trim={0cm 0cm 0cm 0cm},clip,width=0.8\linewidth]{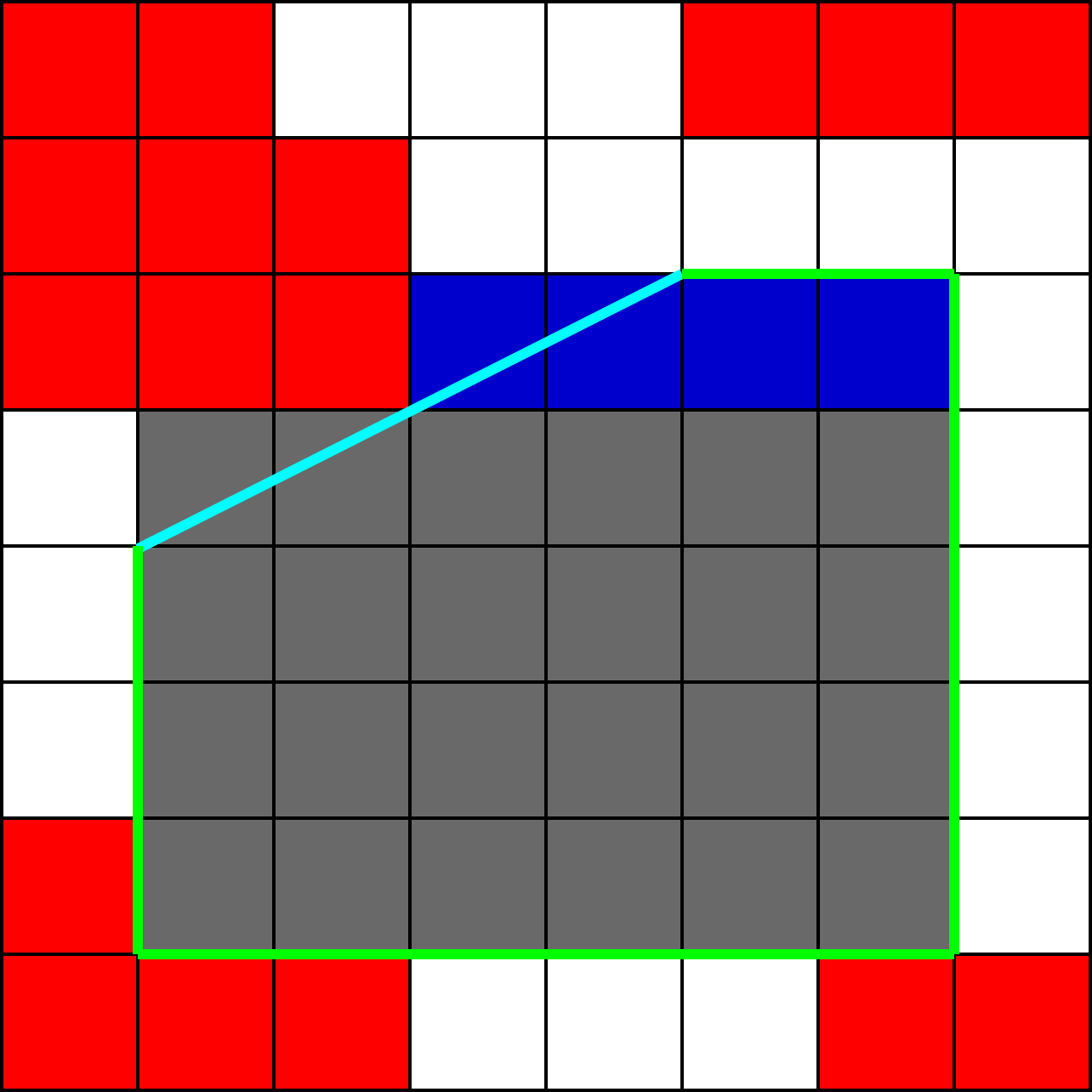}
\caption{Case \textbf{C1}: initial case}
\label{fig:e1_1}
\end{subfigure}
\begin{subfigure}{0.33\textwidth}
\centering
\includegraphics[trim={0cm 0cm 0cm 0cm},clip,width=0.8\linewidth]{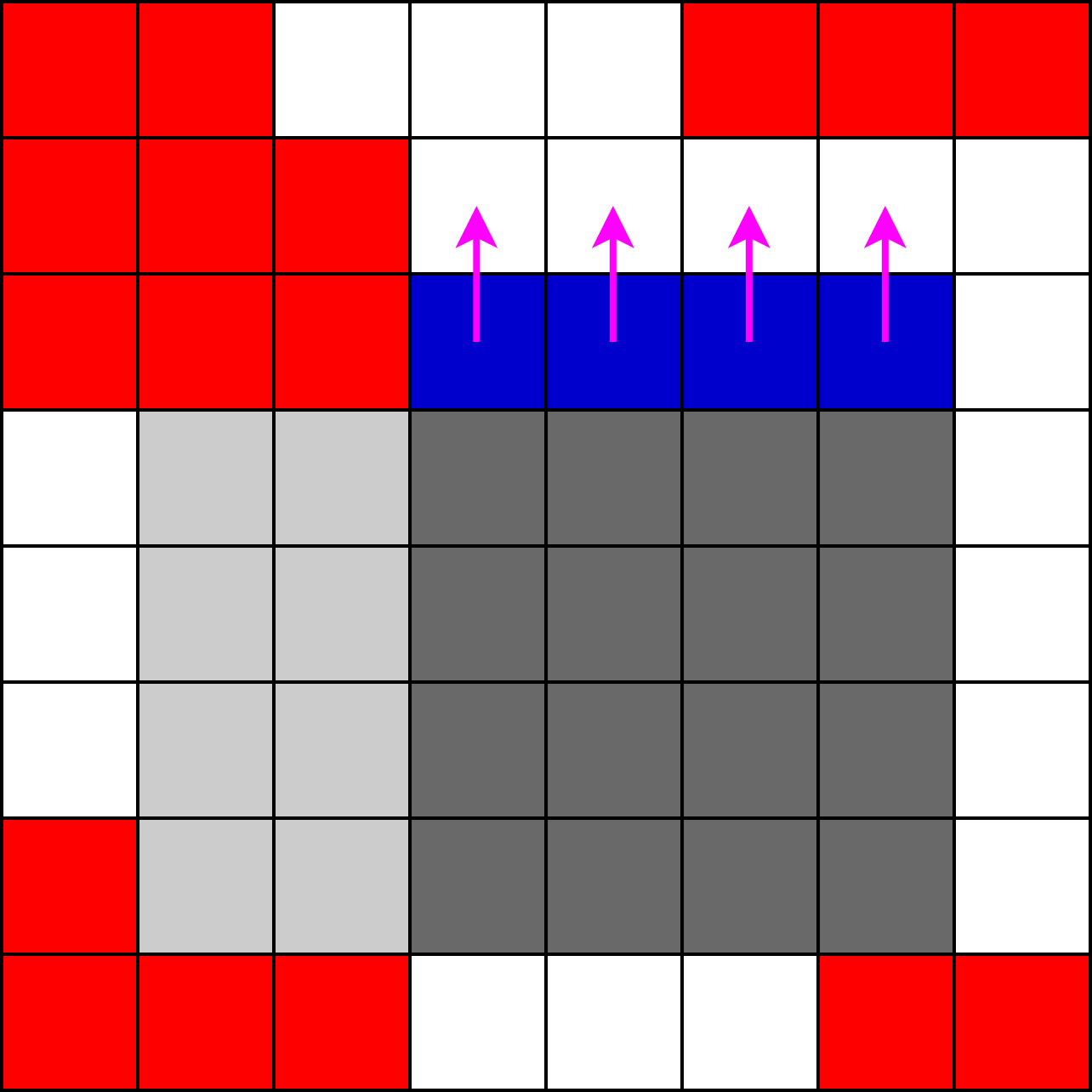}
\caption{Case \textbf{C1}: \textbf{E1} expansion - step 1.}
\label{fig:e1_2}
\end{subfigure}
\begin{subfigure}{0.33\textwidth}
\centering
\includegraphics[trim={0cm 0cm 0cm 0cm},clip,width=0.8\linewidth]{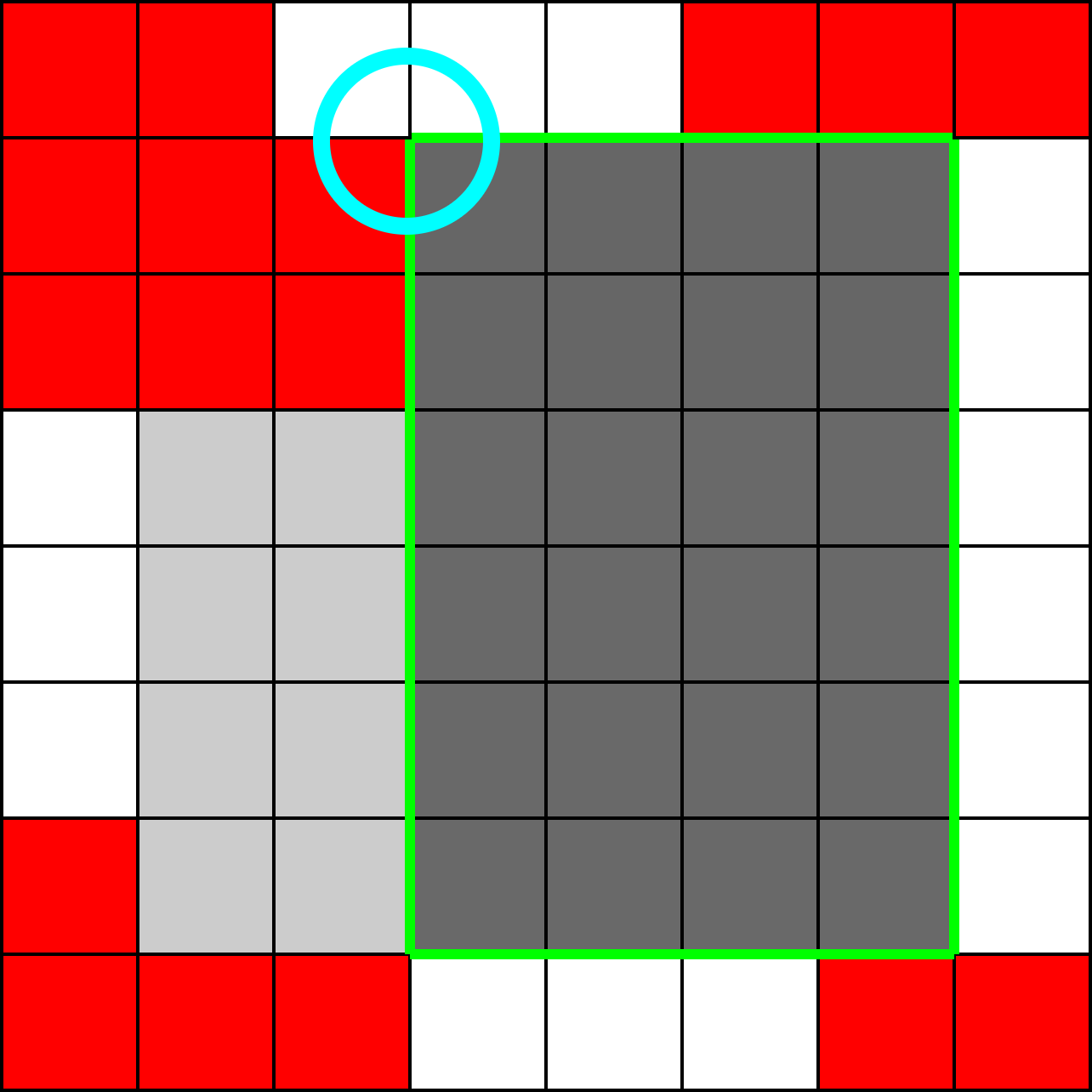}
\caption{Case \textbf{C1}: \textbf{E1} expansion - step 2.}
\label{fig:e1_3}
\end{subfigure}
\caption{We show a 2D example of the case \textbf{C1} and the corresponding expansions (\textbf{E1}). We expand in the same direction that we expanded in to get the new border candidate \ref{fig:e1_2}. This is done by ignoring the cells of the convex gird that are on the side adjacent to the newly initialized corner (depicted in light gray), which is also the same as saying the newly initialized corner does not exist. Once the new convex grid and the corresponding inscribed polyhedron are obtained \ref{fig:e1_3}, we check for the newly initialized corner in the new inscribed polyhedron (encircled in cyan). If it does not exist (which is the case here), the border expansion that resulted in the new border candidate is not valid.}
\label{fig:e1}
\end{figure*}

If we have newly initialized corners, and after the first check if the expansion is still valid, we proceed to the second check which is done for every newly initialized corner: We first expand the new border candidate in the same expansion direction that gave us the new border candidate using the legacy rules (we call this expansion \textbf{E1}). This time the expansion is done by assuming that all the newly initialized corners do not exist, or in other words the cells of the convex grid that are on the side adjacent to the newly initialized corner do not exist \ref{fig:e1_2}. The new expansion will now modify all corners accordingly. 

\begin{figure*}
\begin{subfigure}{0.33\textwidth}
\centering
\includegraphics[trim={0cm 0cm 0cm 0cm},clip,width=0.8\linewidth]{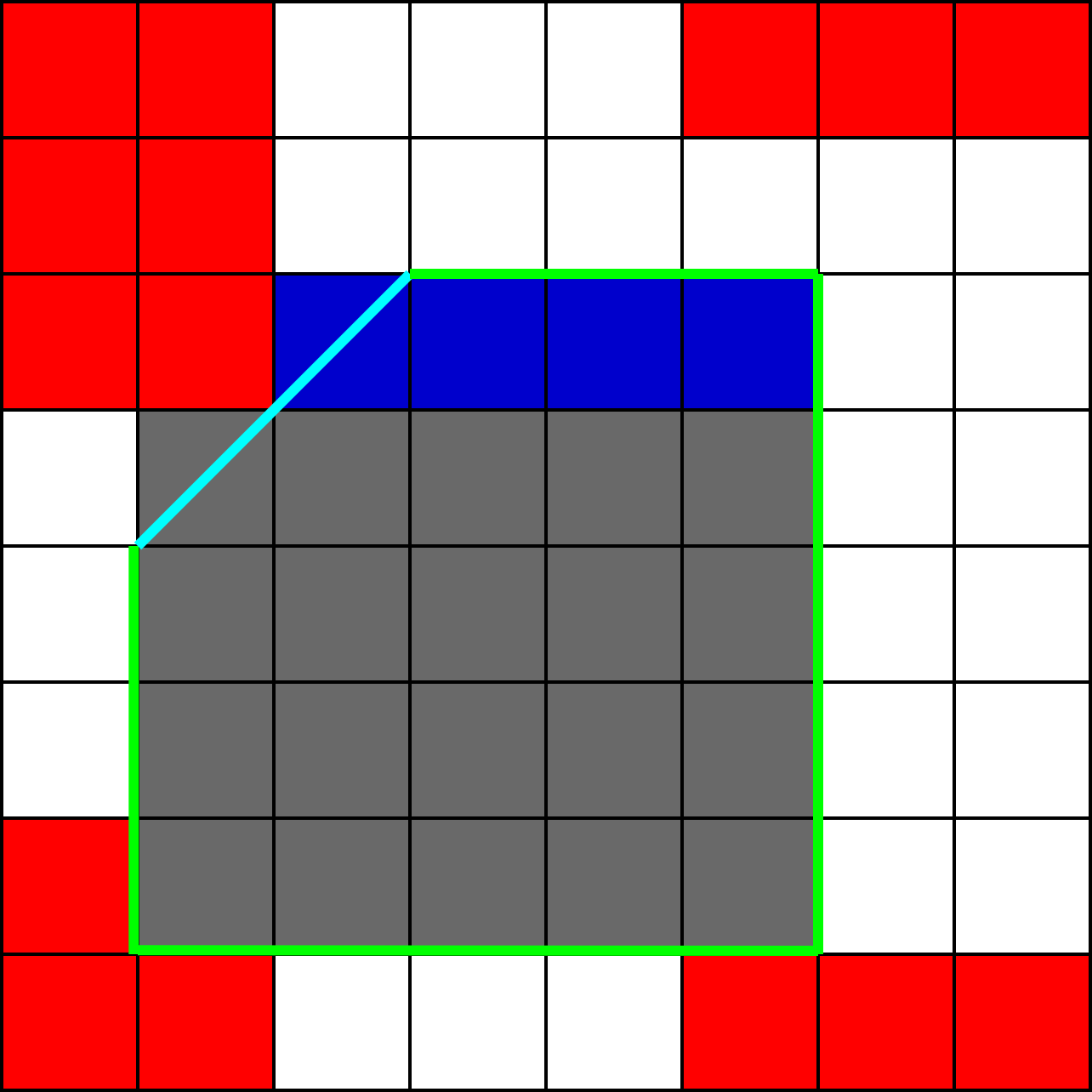}
\caption{Case \textbf{C2}: initial case}
\label{fig:e2_1}
\end{subfigure}
\begin{subfigure}{0.33\textwidth}
\centering
\includegraphics[trim={0cm 0cm 0cm 0cm},clip,width=0.8\linewidth]{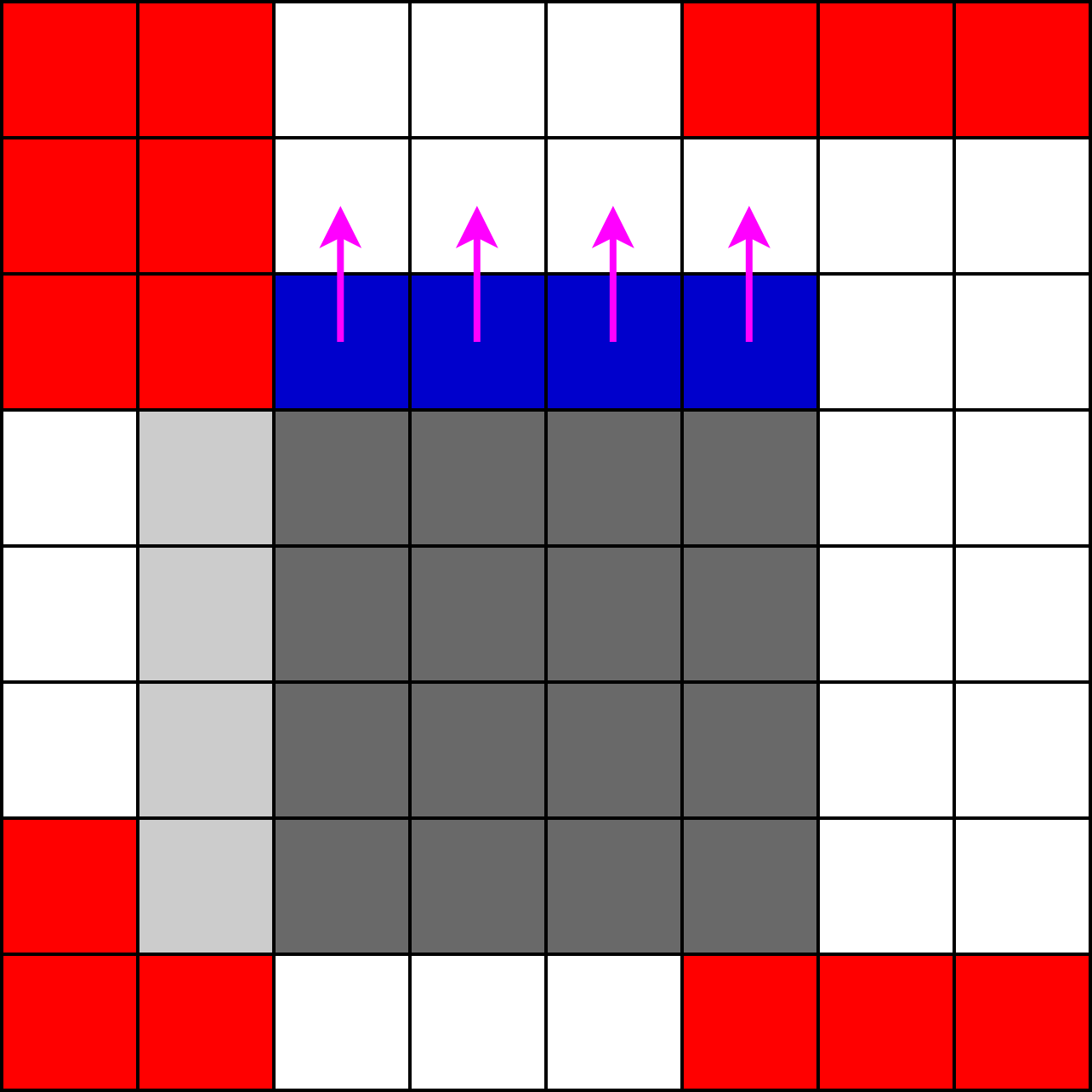}
\caption{Case \textbf{C2}: \textbf{E1} expansion - step 1.}
\label{fig:e2_2}
\end{subfigure}
\begin{subfigure}{0.33\textwidth}
\centering
\includegraphics[trim={0cm 0cm 0cm 0cm},clip,width=0.8\linewidth]{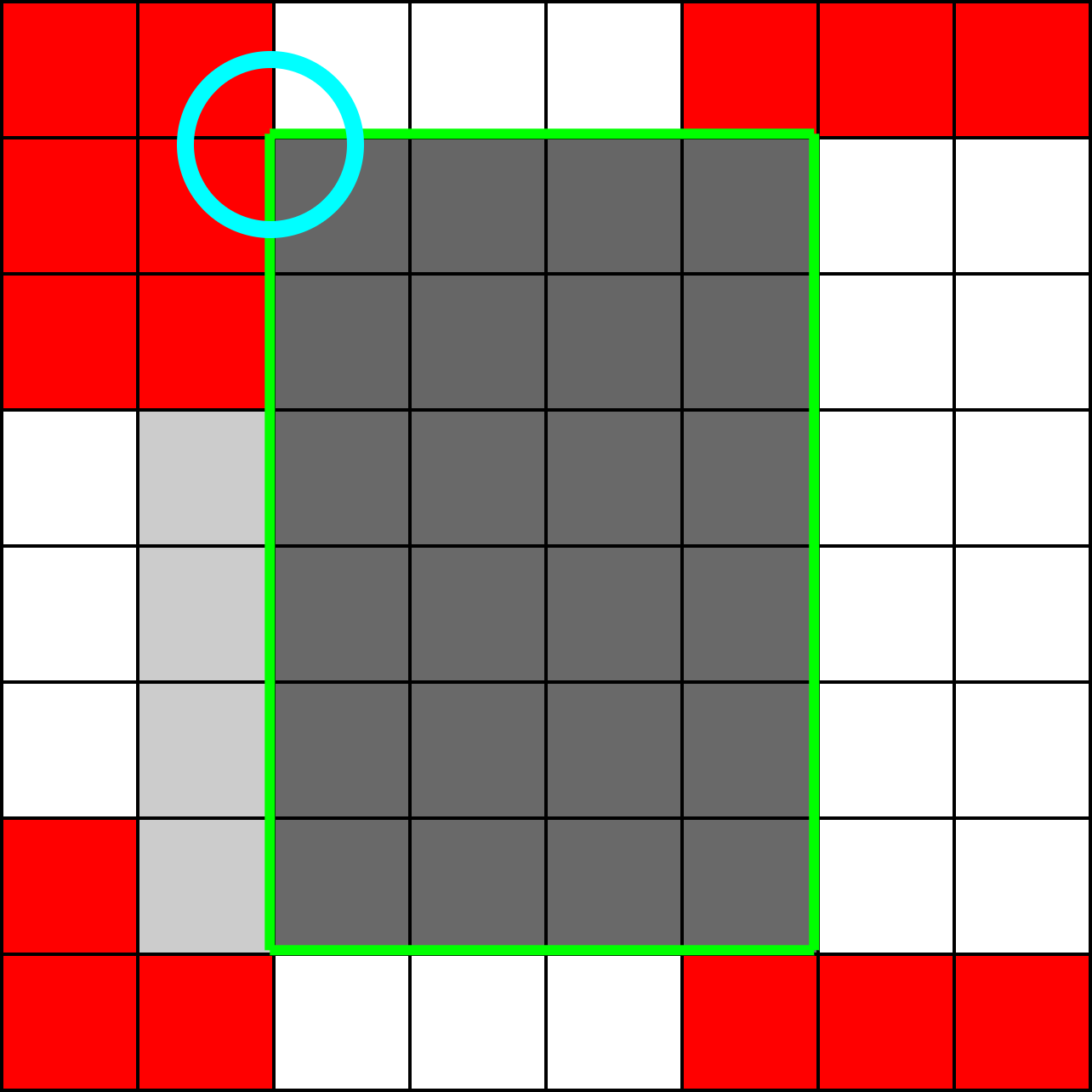}
\caption{Case \textbf{C2}: \textbf{E1} expansion - step 2.}
\label{fig:e2_3}
\end{subfigure}

\begin{subfigure}{0.5\textwidth}
\centering
\includegraphics[trim={0cm 0cm 0cm -0.5cm},clip,width=0.6\linewidth]{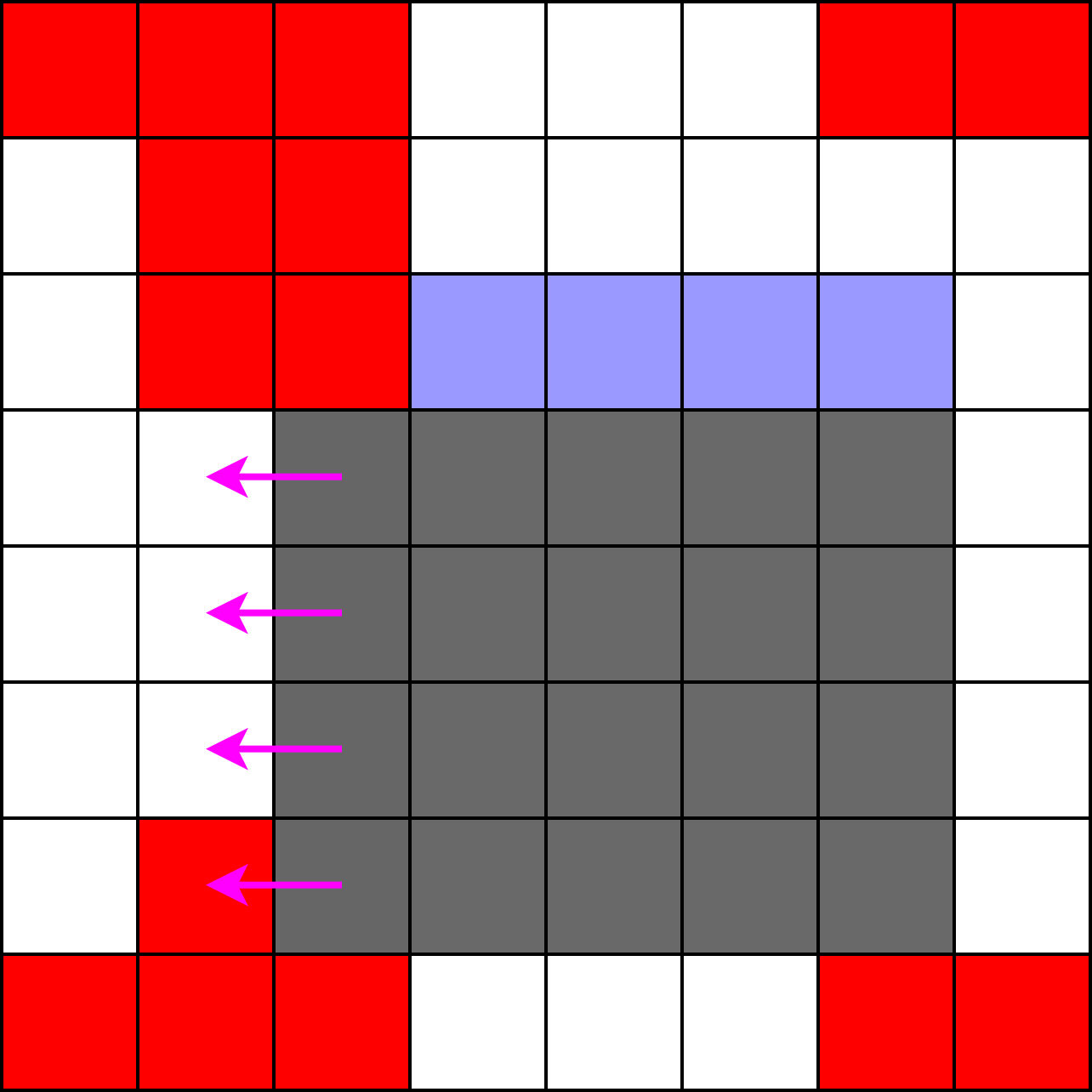}
\caption{Case \textbf{C2}: \textbf{E2} expansion - step 1.}
\label{fig:e2_4}
\end{subfigure}
\begin{subfigure}{0.5\textwidth}
\centering
\includegraphics[trim={0cm 0cm 0cm -0.5cm},clip,width=0.6\linewidth]{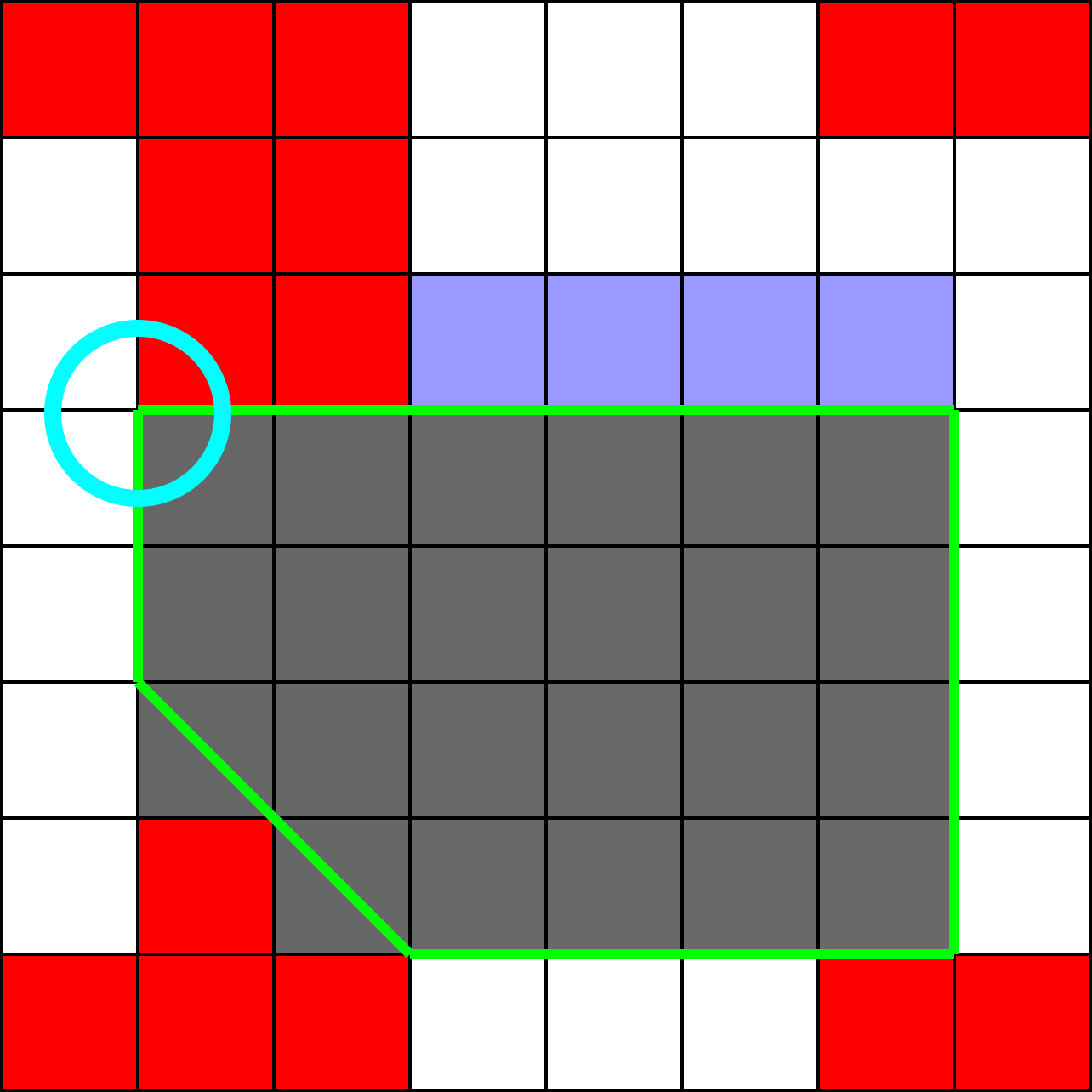}
\caption{Case \textbf{C2}: \textbf{E2} expansion - step 2.}
\label{fig:e2_5}
\end{subfigure}

\caption{We show a 2D example of the case \textbf{C2} and the corresponding expansions (\textbf{E1} and \textbf{E2}). The expansion \textbf{E1} is done as shown in Fig. \ref{fig:e1}. The expansion \textbf{E2} is done by ignoring the new border candidate (depicted in light blue) i.e. assuming the newly initialized corner does not exist, and expanding in the adjacent direction of the newly initialized corner. The new expansion gives us a new value for the newly initialized corner. If by both expansions the newly initialized corner does not exist (which is encircled in cyan in \ref{fig:e2_3} and \ref{fig:e2_5}), the expansion is invalidated.}
\label{fig:e1_e2}
\end{figure*}

If the new expansion is valid according to the legacy rules, and if we are in the case \textbf{C1}, we check if the slope of the corner that we obtain by the new expansion is smaller then the one obtained by the new border candidate \ref{fig:e1_3}. If that is the case, the expansion is invalid. If we are in case \textbf{C2}, we also expand the other border adjacent to the newly initialized corner while assuming the newly initialized corner does not exist yet i.e. we discard the voxels of the new border candidate from the convex grid (we call this expansion \textbf{E2} - Fig. \ref{fig:e2_4}). If both \textbf{E1} and \textbf{E2} result in the newly initialized corner having 0 slope, the expansion is invalid.

Note that these conditions invalidate expansions. In all other cases the expansion is validated (unless invalidated by legacy rules) and the inscribed polyhedron is modified accordingly.

The ability of the proposed method to sense its surroundings before making a decision on whether to expand in a direction or not is what gives it a shape-aware characteristic.

\section{Finding the connectivity graph}
\begin{figure}
\begin{subfigure}{1\linewidth}
\centering
\includegraphics[trim={0cm 0cm 0cm 0cm},clip,width=0.8\linewidth]{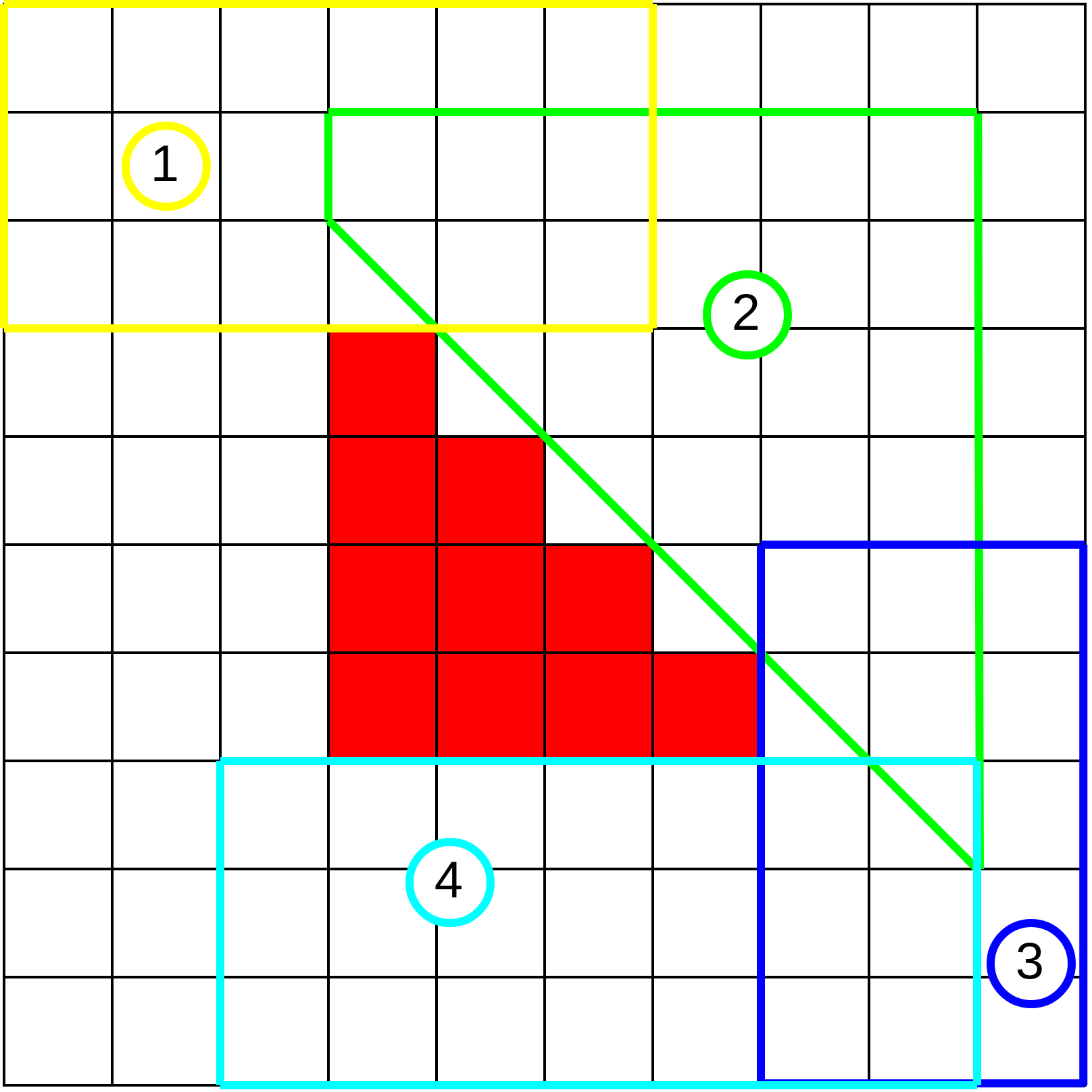}
\caption{Example of multiple polyhedra covering free space and intersecting with each others. Each single polyhedron is colored and numbered differently.}
\label{fig:con_graph_1}

\end{subfigure}
\begin{subfigure}{1\linewidth}
\centering
\includegraphics[trim={0cm 0cm 0cm -0.5cm},clip,width=0.8\linewidth]{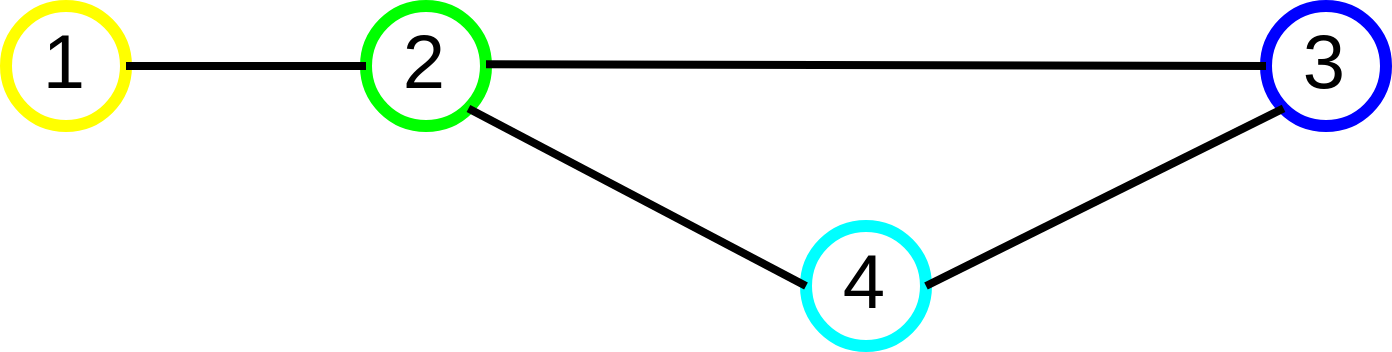}
\caption{The connectivity graph resulting from the polyhedra shown in Fig. \ref{fig:con_graph_1}.}
\label{fig:con_graph_2}
\end{subfigure}
\caption{We show a 2D example of multiple polyhedra covering free space and intersecting with each others in Fig. \ref{fig:con_graph_1} as well as the resulting connectivity graph in Fig. \ref{fig:con_graph_2} that indicates if we are in a given polyhedron, which polyhedra we can move to.}
\label{fig:con_graph}
\end{figure}

Creating a connectivity graph between all the polyhedra that cover space can be beneficial for the planning phase: this would allow us to know which polyhedra we can move to from our current polyhedron i.e. reduce the number of polyhedra in optimization and thus reduce optimization time. An example can be shown in Fig. \ref{fig:con_graph}. Note that the polyhedra in question are not necessarily generated around a path to create a Safe Corridor, but can be generated to cover all the free space within a given area by carefully choosing the seed of each polyhedron.

We use the convex grid/polyhedron duality in order to find whether two polyhedron intersect with each others. This is done by checking if any of the voxels in the convex grid of one polyhedron also belongs to the convex grid of another polyhedron.  

If we use this method to find if an intersection exists between 2 polyhedra and their corresponding convex grids that contain $216$ voxels, the computation time will be on average $2 \ \mu s$ (on an Intel Core i7-9750H up to 4.50 GHz CPU). Other state of the art methods \cite{tordesillas2020separator} \cite{tordesillas2020mader} try to find a separating plane between the 2 polyhedra. If that plane exists then the polyhedra do not intersect. This method takes on average $200 \ \mu s$ i.e. $100 \times$ slower then the method proposed in this paper.

In the example shown in Fig. \ref{fig:con_graph}, we need to check each polyhedra with all the others to see if there is an intersection and create the connectivity graph shown in Fig. \ref{fig:con_graph_2}. This means we need to check $3 + 2 + 1 = 6$ times. Using our method would take $12 \ \mu s$ whereas using \cite{tordesillas2020separator} would take $1.2\ ms$.

\section{Simulation results}
\subsection{Simulation setup}
The simulation is run on an Intel® Core™ i7-9750H (base 2.60GHz, up to 4.50 GHz).
  All distances and coordinates are in meters. The size of the environment is $50\times 12\times 12$. We represent the environment as a voxel grid \cite{toumieh2020mapping} whose voxel size is $0.3$. $400$ obstacles are generated, of size ranging from $0.9$ to $1.5$ (following a uniform distribution). The obstacle is contained in a cube (which we call \textbf{obstacle cube}) with side length between 3 ($0.9/voxel\_size$)and 5 ($1.5/voxel\_size$) voxels. We randomize the shape of the obstacle by setting each voxel inside the \textbf{obstacle cube} to occupied with a probability of $0.5$. The obstacles position is generated using a uniform distribution.

The starting point is set to $(x=3,y=6,z=6)$ and the goal is set to $(x=47,y=6,z=6)$. The JPS and DMP take on average $6 \ ms$ (combined). \cite{liu2017planning} is optimized by using the occupied voxels of the voxel grid to get the obstacle points that are within a bounding box distance from the seed. This bounding box is set to $4\times4\times4$. The number of border expansions is set to $n = 36$ for \cite{toumieh2020convex} and for the proposed method in this paper in order to cover the same volume ($6$ border expansions to do a full rotation and expand by the voxel size in every direction - Fig. \ref{fig:seed_decomp}).

\subsection{Simulation results}

\begin{table*}[ht]
\centering
\caption{Comparison between Liu et al. \cite{liu2017planning}, Toumieh et al. \cite{toumieh2020convex} and the method proposed in this paper on 10 randomly generated maps of size $50\times 12\times 12$ and with 400 obstacles. The \textbf{mean / max / standard deviation} of every metric is shown. The difference in performance between Toumieh et al. \cite{toumieh2020convex} and the proposed method is shown for the \textbf{mean} and \textbf{max} values. The better performer between the 3 methods is shown in bold.}
\begin{tabular}{c | ccccc}\hline
 & Volume (m\textsuperscript{3}) & Constr/Poly & Poly/SC & Comp. time ($\mu s$) & Safe \\ \hhline{======}
Liu et al. \cite{liu2017planning} & \textbf{482} / \textbf{519} / 29.5 & 15.2 / 26 / 3.2 & 27.5 / 33 / 2.8 & \textbf{107} / \textbf{222} / 37.7 & no \\ \hline
Toumieh et al. \cite{toumieh2020convex} & 414 / 456 / 29.8 & 10.9 / 16 / 1.7 & 27.8 / \textbf{31} / 2.1 & 187 / 349 / 54 & \textbf{yes} \\ \hline
Proposed method & 399 / 446 / 35 & \textbf{7} / \textbf{12} / 1 & \textbf{27.3} / \textbf{31} / 2 & 191 / 337 / 52 & \textbf{yes} \\ \hhline{======}
Difference (\%) & -3.6/-2.4 & -35.7/-25 & -1.8/0 & +2/-3.4 & -
\end{tabular}
\label{tab:comp}
\end{table*}

The proposed method in this paper is compared with the method proposed in \cite{liu2017planning} as well as \cite{toumieh2020convex} in finding a Safe Corridor (SC) between the same starting and goal points in 10 randomly generated environments. The comparison metrics are volume covered, number of constraints per polyhedron, number of polyhedra per SC (genericness), computation time and SC safety. The mean, max, and standard deviation of the relevant metrics are shown in Tab. \ref{tab:comp}. The difference in the metrics between the 2 safe methods i.e. \cite{toumieh2020convex} and our method, is also shown in Tab. \ref{tab:comp}.

In terms of volume covered, Liu et al. \cite{liu2017planning} is the best performer, but this is in part due to the lack of safety: in fact, the polyhedra generated by \cite{liu2017planning} can penetrate between the obstacle points (Fig. \ref{fig:liu_method}) and thus have non empty intersection with the real world obstacles since the obstacle voxel points are a downsampled version of the pointcloud that represents the real obstacles. Our proposed method generates Safe Corridors that cover on average $3.6 \%$ less volume then Toumieh et al. \cite{toumieh2020convex}.

In terms of constraints per polyhedron, our method is the best performer with an average of 7 constraints per polyhedron, $35.7 \%$ smaller then the next best method Toumieh et al. \cite{toumieh2020convex} (which has on average 10.9 constraints per polyhedron). Note that a small number of constraints is favourable for planning since it results in smaller computation times \cite{toumieh2020planning}.

All 3 methods are similar in terms of number of polyhedra per Safe Corridor with the difference in the average being less then $2 \%$.

Finally in terms of computation time, Liu et al. \cite{liu2017planning} is the best performer being approximately $1.8\times$ faster than Toumieh et al. \cite{toumieh2020convex} and our method, who have similar computation times. However both Toumieh et al. \cite{toumieh2020convex} and our method remain largely within the real-time constraints necessary for planning as shown in \cite{toumieh2020planning}. 

\subsection{Case study}
\begin{figure*}
\begin{subfigure}{1\textwidth}
\centering
\includegraphics[trim={0cm 0cm 0cm 0cm},clip,width=0.9\linewidth]{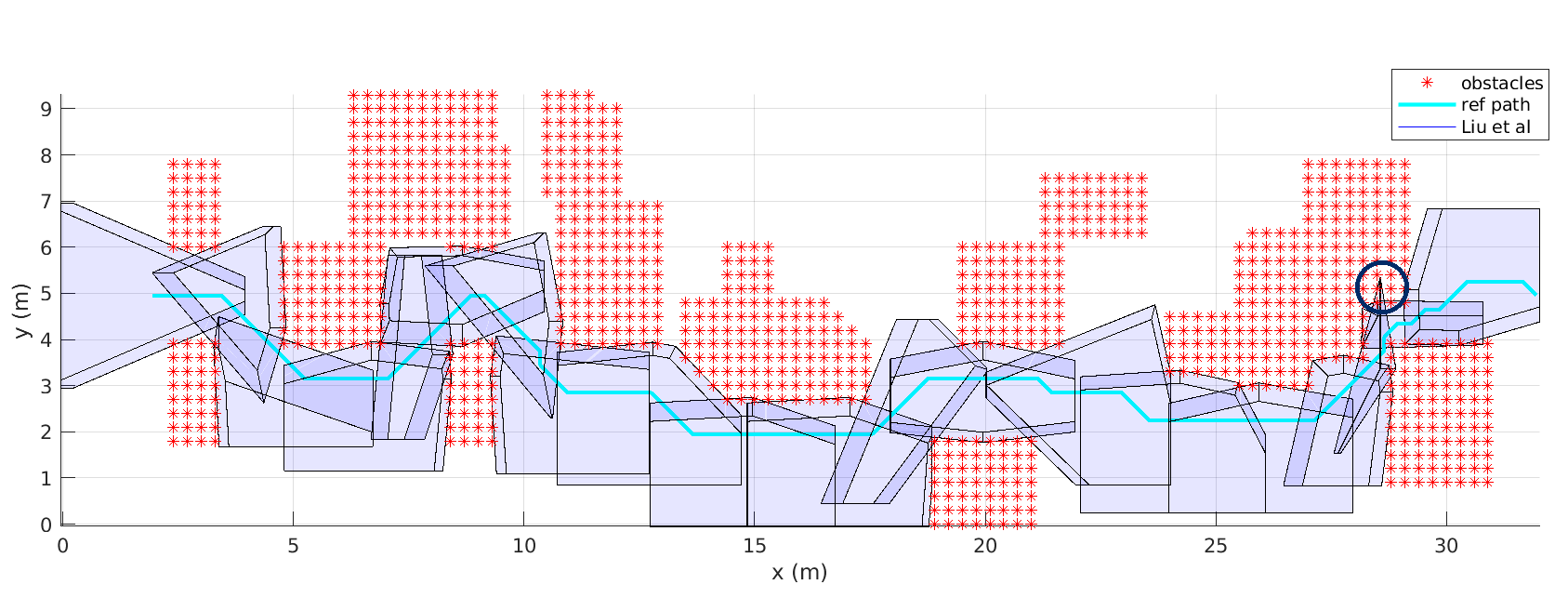}
\caption{Liu et al's method \cite{liu2017planning}}
\label{fig:liu_method}
\end{subfigure}

\begin{subfigure}{1\textwidth}
\centering
\includegraphics[trim={0cm 0cm 0cm 0cm},clip,width=0.9\linewidth]{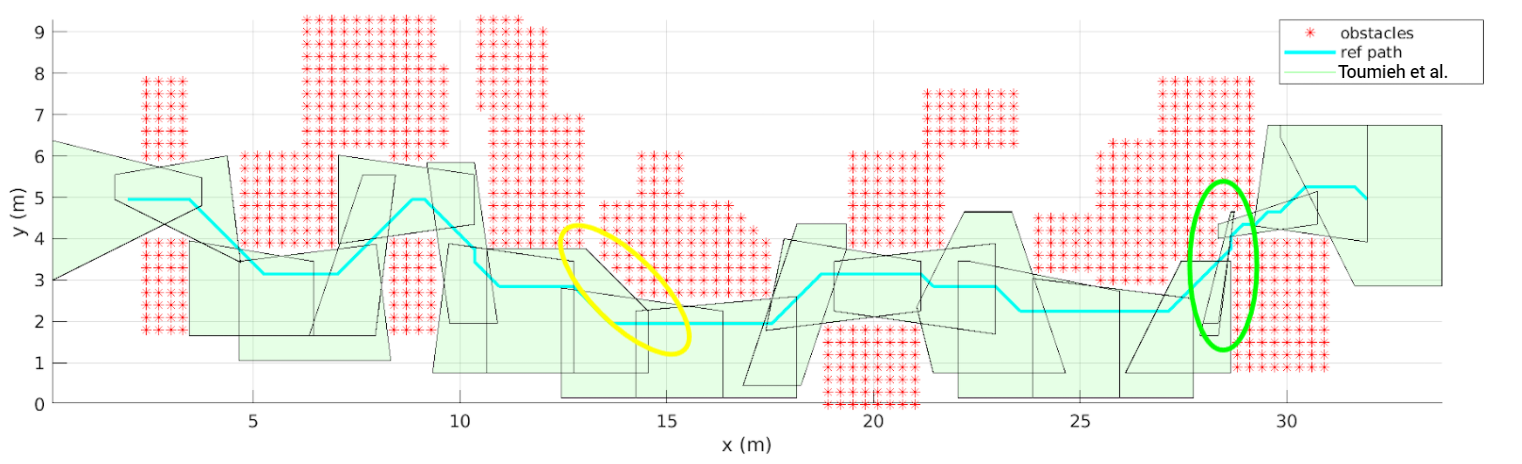}
\caption{Toumieh et al.'s method \cite{toumieh2020convex}}
\label{fig:toumieh_method}
\end{subfigure}

\begin{subfigure}{1\textwidth}
\centering
\includegraphics[trim={0cm 0cm 0cm 0cm},clip,width=0.9\linewidth]{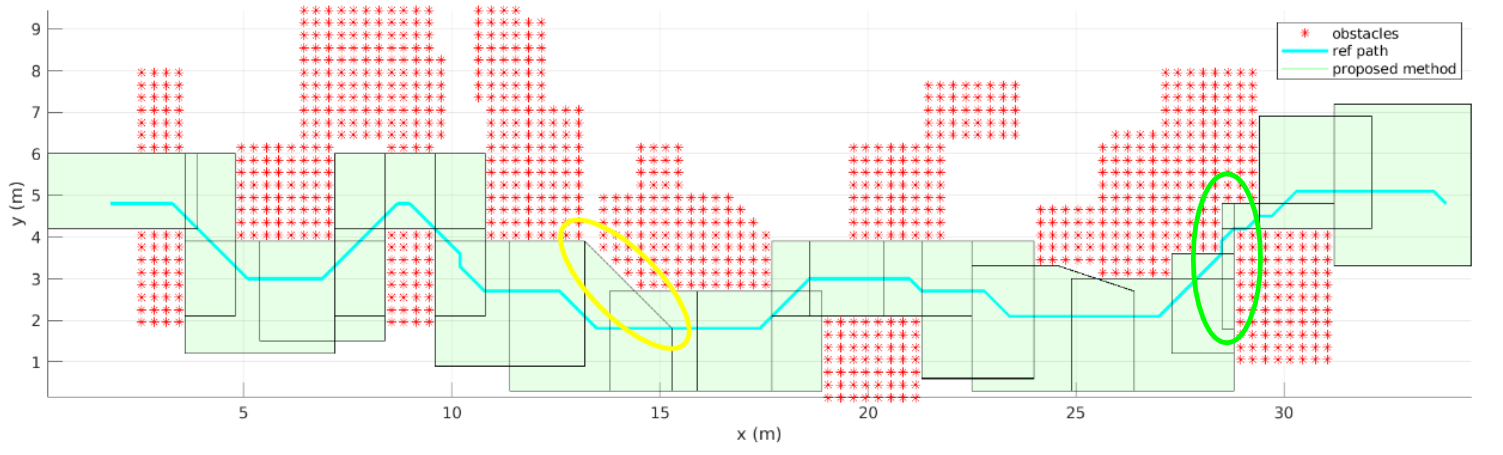}
\caption{Proposed method}
\label{fig:our_method}
\end{subfigure}

\caption{An overhead view of the Safe Corridor generated around a path using Liu et al. \cite{liu2017planning}, Toumieh et al. \cite{toumieh2020convex} and the proposed method in this paper. The place where a polyhedron generated by Liu et al.'s method penetrates the space between the obstacle points of the downsampled point cloud is encircled in \textbf{dark blue}. We encircled in green a place where our method generates a higher quality decomposition then the other safe method (Toumieh et al.). We encircled in yellow a place where a slope in the obstacles is captured by our method and Toumieh et al.'s method to show that the improvement shown in the green ellipse doesn't hinder the genericness of our method in covering space efficiently around sloped obstacles (encircled in yellow).}
\label{fig:decomp}
\end{figure*}

While the aforementioned performance metrics provide some insight into the comparative performance of these methods, we show in Fig. \ref{fig:decomp} an overhead view of the decomposition of a particular environment to better emphasize the differences. The obstacle points shown are the centers of the occupied voxels in a voxel grid. The overhead view shows a horizontal slice (in the $x-y$ plane) of the obstacles i.e. the shown obstacle points are repeated at every voxel level in the $z$ direction to create a 3 dimensional obstacle.

As shown in \cite{toumieh2020convex}, the method by Liu et al. \cite{liu2017planning} generates unsafe corridors. We reproduce the results shown in \cite{toumieh2020convex} in Fig. \ref{fig:liu_method} and encircle in blue a place where the polyhedron of the Safe Corridor penetrates between the obstacle points and thus intersect with the real world obstacles. This is due to the nature of the algorithm they use (as described in \ref{sect:related_work}) and to the fact that we use the voxel grid to downsample the pointcloud to accelerate the convex decomposition as done in \cite{tordesillas2020faster}. 

In Fig. \ref{fig:toumieh_method} and Fig. \ref{fig:our_method} we compare our method directly to Toumieh et al. \cite{toumieh2020convex}. We encircled in green and yellow the areas of interest that showcase the difference/similarity between the 2 methods. 

First we encircled in green a part of the Safe Corridor where our method performs clearly better due to the fact that it is shape-aware and it can sense the surroundings before creating a polyhedron. In the case of Toumieh et al. \cite{toumieh2020convex} (Fig. \ref{fig:toumieh_method}), one can see that in this tight corner (encircled in green), there are still free spaces that are not covered by the polyhedra of the Safe Corridor. Whereas in our case, the decomposition covers the free space perfectly, due to the fact that the algorithm recognizes that there is a sharp corner and doesn't generate a polyhedron with faces not parallel to the $x$ and $y$ directions.

In the area encircled in yellow, we show how Toumieh et al. \cite{toumieh2020convex} and our method perform in the presence of obstacles that have faces that are not parallel to the $x$ or $y$ directions i.e. \textit{sloped} faces. The ability to cover free space around obstacles of arbitrary \textit{slopes}/shapes is a measure of genericness of the method. Both methods succeed in capturing the slope in the polyhedron close to it. This is to show that while our method performs well in environments with rectangular shaped obstacles, it also performs just as well as Toumieh et al. \cite{toumieh2020convex} when the obstacles have \textit{sloped} faces.

\section{Conclusions and future work}
In this paper, we presented a new framework for creating Safe Corridors and finding the connectivity graph between multiple polyhedra. The generation of the connectivity graph is highly computationally efficient (due to using the convex grid/polyhedron duality), and can be used in the planning phase to know which polyhedra can the robot go to if he is in a given polyhedron. The Safe Corridor generation is an extension of the work done in \cite{toumieh2020convex}: We added a few conditions to make the method in \cite{toumieh2020convex} sense its surroundings before making a decision on the shape of each polyhedron. This gives our method a shape-aware characteristic, and allows it to cover the free space around obstacles more efficiently. We compared our method to other state-of-the-art methods \cite{liu2017planning} \cite{toumieh2020convex} in simulations quantitatively and qualitatively and showed that it performs better in terms of quality. 

% \hl{We also plan to investigate how to add dynamic obstacles in the planning framework, more specifically the MIQP/MPC formulation.} %ajouter référence MPC
% arreter ici , cela est déjà suffisant comme suite à ces travaux.. %% done


%%%%%%%%%%%%%%%%%%%%%%%%%%%%%%%%%%%%%%%%%%%%%%%%%%%%%%%%%%%%%%%%%%%%%%%%%%%%%%%%

%%%%%%%%%%%%%%%%%%%%%%%%%%%%%%%%%%%%%%%%%%%%%%%%%%%%%%%%%%%%%%%%%%%%%%%%%%%%%%%%

%%%%%%%%%%%%%%%%%%%%%%%%%%%%%%%%%%%%%%%%%%%%%%%%%%%%%%%%%%%%%%%%%%%%%%

\bibliographystyle{IEEEtran}
\bibliography{IEEEabrv,IEEEexample}

\end{document}